\documentclass[sigconf]{acmart}
\usepackage{subfigure}
\usepackage[linesnumbered,ruled,noend]{algorithm2e} 
\usepackage{multirow}
\usepackage{rotating}

\usepackage{multicol}

\copyrightyear{2024}
\acmYear{2024}
\setcopyright{rightsretained}
\acmConference[KDD '24]{Proceedings of the 30th ACM SIGKDD Conference on Knowledge Discovery and Data Mining}{August 25--29, 2024}{Barcelona, Spain}
\acmBooktitle{Proceedings of the 30th ACM SIGKDD Conference on Knowledge Discovery and Data Mining (KDD '24), August 25--29, 2024, Barcelona, Spain}\acmDOI{10.1145/3637528.3671969}
\acmISBN{979-8-4007-0490-1/24/08}

\usepackage{balance}

\begin{document}

\title{Addressing Prediction Delays in Time Series Forecasting: \\A Continuous GRU Approach with Derivative Regularization}

\author{Sheo Yon Jhin}
\affiliation{%
  \institution{Yonsei University}
  \city{Seoul}
  \country{South Korea}
}\email{sheoyonj@yonsei.ac.kr}

\author{Seojin Kim}
\affiliation{%
  \institution{Yonsei University}
  \city{Seoul}
  \country{South Korea}}
\email{bwnebs1@yonsei.ac.kr}

\author{Noseong Park}
\affiliation{%
  \institution{KAIST}
  \city{Daejeon}
  \country{South Korea}
}
\email{noseong@kaist.ac.kr}

\renewcommand{\shortauthors}{Sheo Yon Jhin, Seojin Kim, and Noseong Park}
\begin{abstract}
  Time series forecasting has been an essential field in many different application areas, including economic analysis, meteorology, and so forth. The majority of time series forecasting models are trained using the mean squared error (MSE). However, this training based on MSE causes a limitation known as \textbf{\textit{prediction delay}}. The prediction delay, which implies the ground-truth precedes the prediction, can cause serious problems in a variety of fields, e.g., finance and weather forecasting --- as a matter of fact, predictions succeeding ground-truth observations are not practically meaningful although their MSEs can be low. This paper proposes a new perspective on traditional time series forecasting tasks and introduces a new solution to mitigate the prediction delay. We introduce a continuous-time gated recurrent unit (GRU) based on the neural ordinary differential equation (NODE) which can supervise explicit time-derivatives. We generalize the GRU architecture in a continuous-time manner and minimize the prediction delay through our time-derivative regularization. Our method outperforms in metrics such as MSE, Dynamic Time Warping (DTW) and Time Distortion Index (TDI). In addition, we demonstrate the low prediction delay of our method in a variety of datasets.
\end{abstract}

\ccsdesc[500]{Computing methodologies~Machine learning algorithms}
\ccsdesc[500]{Computing methodologies~Artificial intelligence}

\keywords{Time-series forecasting, Prediction delay, Neural ODE}


\received{20 February 2007}
\received[revised]{12 March 2009}
\received[accepted]{5 June 2009}

\maketitle

\section{Introduction}
\begin{figure}[ht!]
\vskip 0.2in
\begin{center}
\subfigure[Prediction results from PatchTST, DLinear]{\includegraphics[width=1.0\columnwidth]{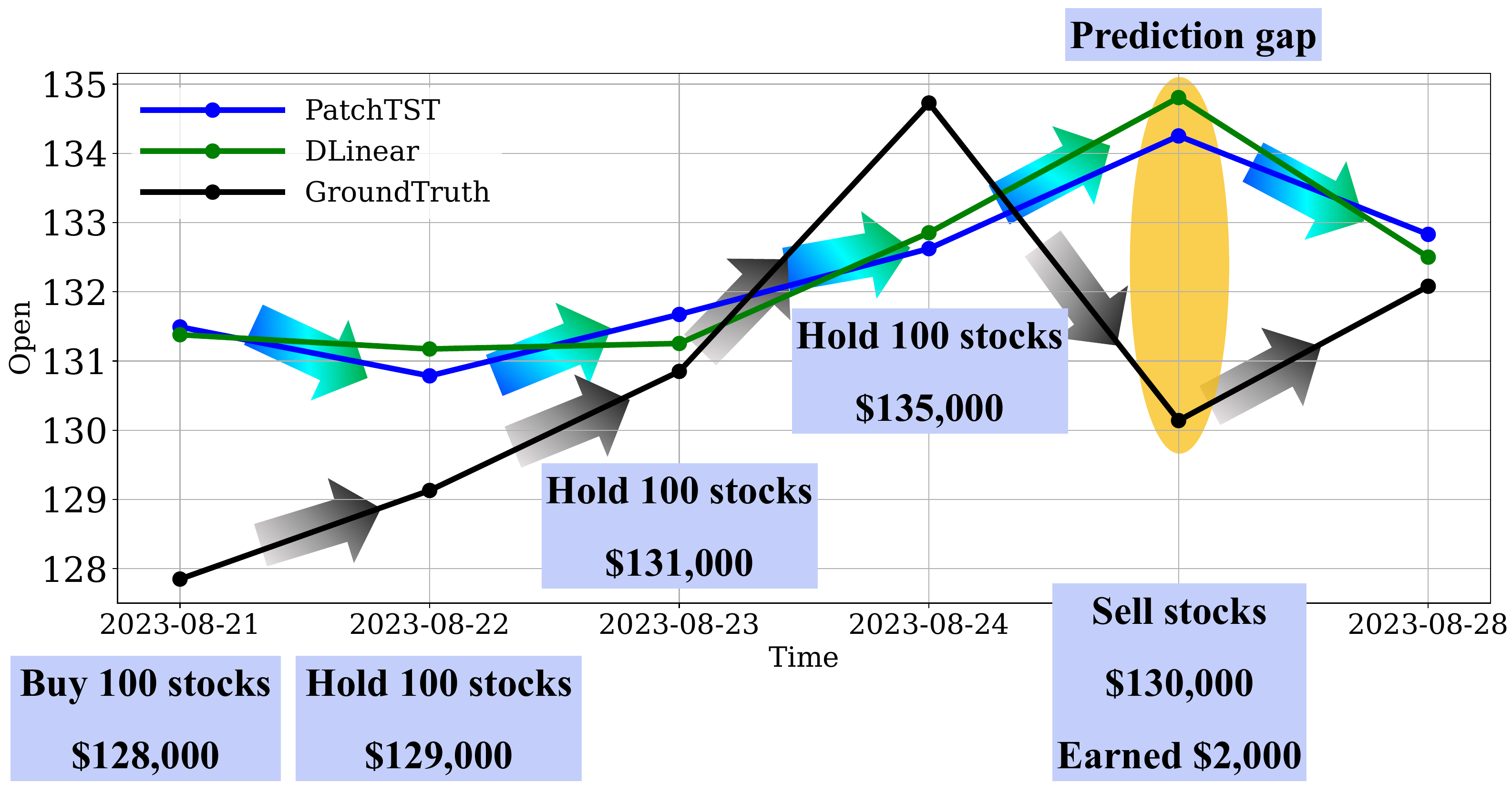}} \\
\subfigure[Prediction results from CONTIME]{\includegraphics[width=1.0\columnwidth]{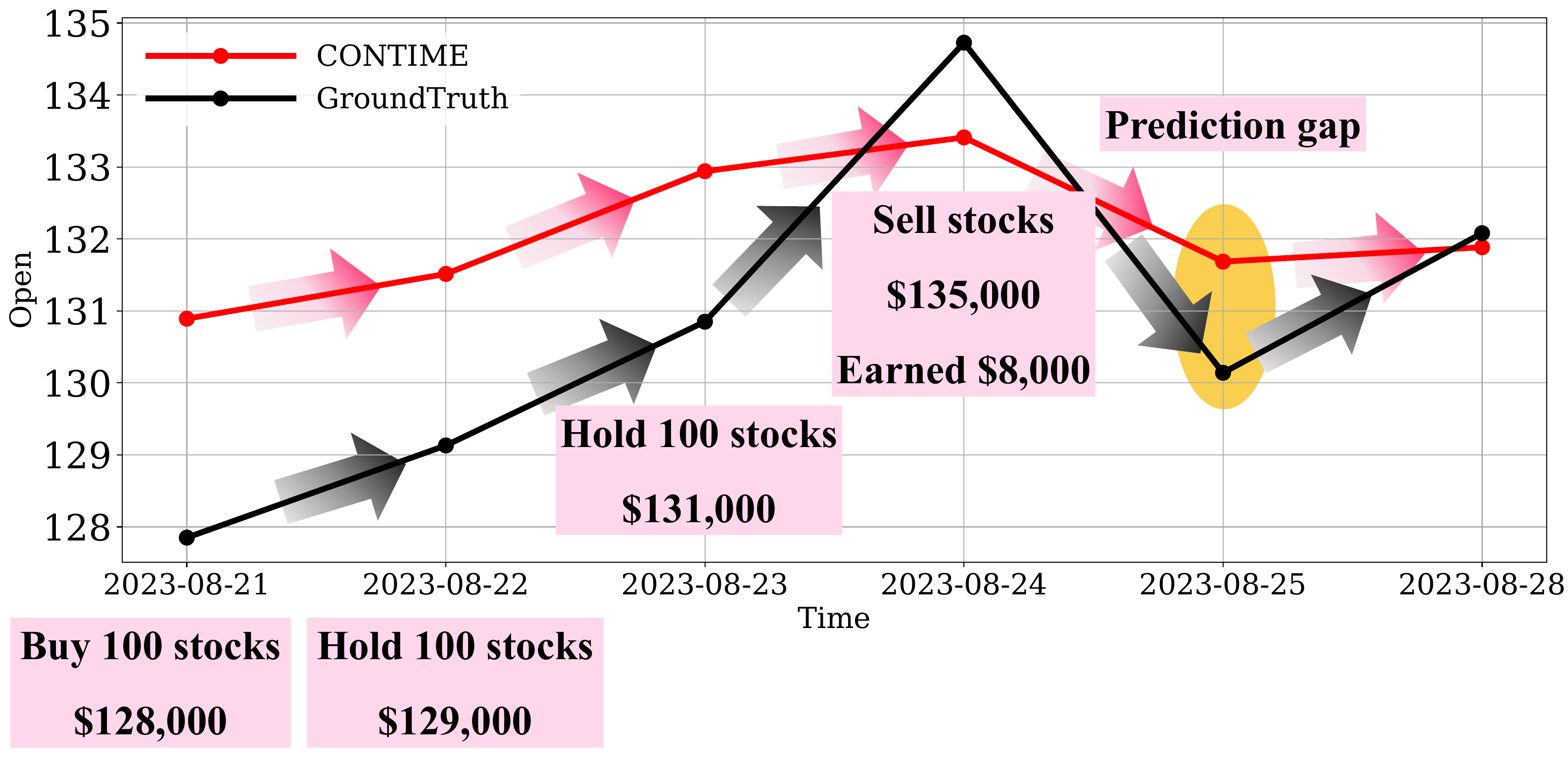}} 
\caption{
Visualization of Table~\ref{tbl:teaser_google} (experimental results for GOOG stock prediction from August 21 to August 28, 2023)}

\label{fig:teaser_img}
\end{center}
\end{figure}

Time series forecasting is important in diverse domains, such as weather prediction, stock prediction, and so forth, and has several challenges~\citep{jhin2022exit,cao2019financial,lim2021time,salman2015weather}. The necessity to tackle these practical challenges has spurred numerous proposed studies investigating the intricacies of short- and long-term time series forecasting. Within this realm, a spectrum of models has been suggested, ranging from simple linear networks to advanced transformer-based architectures~\citep{zhou2021informer,liu2021pyraformer,kitaev2020reformer}. Traditionally, the predominant evaluation metrics in most studies have been Mean Squared Error (MSE) or Mean Absolute Error (MAE), with ongoing research endeavors striving to showcase state-of-the-art outcomes through learning based on these metrics. However, the remarkable success achieved in time series forecasting using MSE highlights a limitation related to the prediction delay, as illustrated in Figure~\ref{fig:teaser_img}.(a). In this context, we define \textbf{\textit{prediction delay}} as a phenomenon where the actual observations precede the prediction in the time series forecasting task --- in other words, a model is trained to output an observation similar to the most recent observation, which can lead to reasonable MSE or MAE values but is, in practice, rather meaningless~\citep{conway1998delayed,dixit2015removing,han2023economic,le2019shape}.
\begin{table}[h!]
    \centering
    \small
    \caption{Experimental results on GOOG on $P=36$}
    \begin{tabular}{cccc} \toprule
        Models   & TDI $\downarrow$  & DTW $\downarrow$  & MSE  $\downarrow$ \\\toprule
        DLinear  & \underline{4.835} & \underline{2.229} & 0.199 \\
        PatchTST & 4.882 & 2.766 & \underline{0.191}\\\midrule
        CONTIME  & \textbf{4.712} & \textbf{2.189} & \textbf{0.189}\\ \bottomrule
    \end{tabular}
    \label{tbl:teaser_google}
\end{table}


Figure~\ref{fig:teaser_img} visually presents the experimental results detailed in Table~\ref{tbl:teaser_google}. Notably, a prediction delay is discernible in state-of-the-art (SOTA) models, exemplified by PatchTST and DLinear~\citep{Yuqietal-2023-PatchTST, zeng2023transformers}, despite their relatively small Mean Squared Error (MSE). Conversely, CONTIME, characterized by a comparatively similar MSE, does not exhibit a prediction delay. This study seeks to provide a comprehensive interpretation of time series forecasting by introducing additional metrics, namely Temporal Distortion Index (TDI) and Dynamic Time Warping (DTW), aimed at elucidating the observed phenomenon. Furthermore, Figure~\ref{fig:teaser_img} underscores the significance of prediction delay in a straightforward scenario. Investors relying on SOTA model forecasts for GOOG stocks anticipate the upper price limit on August 25th, 2023. However, due to a one-day delay in the forecast results, this leads to stock sales (See Figure~\ref{fig:teaser_img}.(a)). In contrast, investors relying on CONTIME, free from prediction delay, predict a stock price decline on August 25th, 2023, prompting them to initiate stock sales on August 24th, 2023 (See Figure~\ref{fig:teaser_img}.(b)). Assuming an investor trades 100 shares of stock, those relying on CONTIME stand to make a profit of approximately \$6,000. This achieves more accurate and timely forecasts in real-world applications and provides beneficial forecast results to investors. The example in Figure~\ref{fig:teaser_img} highlights the importance of mitigating the prediction delay in time series forecasting.

Beyond the financial domain, the aforementioned prediction delay assumes a significant role in areas intricately interwoven with daily life, such as weather forecasting. Despite discussions on these limitations dating back to 1998~\citep{conway1998delayed,de2005constraints,abrahart2007timing,dixit2015removing,han2023economic,le2019shape}, recent state-of-the-art studies have predominantly concentrated on evaluating the performance of metrics like MSE, MAE, etc., in the context of time series forecasting. As evident from \cite{le2019shape}, it is imperative that a model's prediction accurately captures both the \textit{shape} and \textit{temporal} trends within the time series. Dynamic Time Warping (DTW) emerges as a method capable of discerning differences in shape between time series. Additionally, Temporal Distortion Index (TDI) serves as an extra metric to explore the temporal lag between two sequences. The incorporation of these two metrics enables a comprehensive assessment of comparability between the respective time series. Consequently, our intention is to subject the model to evaluation using these novel metrics.



This paper introduces an innovative approach to mitigate the prediction delay in time series forecasting. In this paper, we redefine GRU as differential equation that reflect past observations to the current hidden state for processing continuously generalizing GRU. We propose a continuous-time bi-directional gated recurrent unit (GRU) network based on neural ordinary differential equation (NODE) and train it with explicit time-derivative regularizations, thus addressing the inherent prediction delay observed in various time series forecasting models. We extend the bi-directional GRU to efficiently capture the temporal dependencies within time-series sequences with minimal delays. Our contributions can be summarized as follows:

\begin{enumerate}
    \item We propose \textbf{CON}tinuous GRU to address the prediction delay in \textbf{TIME} series forecasting, i.e., \textbf{CONTIME}. By continuously extending the bi-directional GRU, we present a novel architecture that facilitates the supervision of the time-derivative of observations in the continuous time domain.
    \item In Section~\ref{sec:bi_d_contime}, we compute the time-derivatives of the hidden state $\mathbf{h}(t)$, the reset gate $\mathbf{r}(t)$, the update gate $\mathbf{z}(t)$, and the update vector $\mathbf{g}(t)$ of GRU. We strategically employ the bi-directional GRU structure to generate more effective hidden representations for downstream task.
    \item We conduct time series forecasting with minimal prediction delays through our proposed time-derivative regularization.
    \item \textbf{CONTIME} demonstrates outstanding performance in addressing the prediction delay across all 6 benchmark datasets. In addition to minimizing the prediction delay, it excels in all three metrics (TDI, DTW, and MSE).
    \item Our code is available at this link \footnote{\url{https://github.com/sheoyon-jhin/CONTIME}}, and we refer readers to Appendix~\ref{appendix:hyperparameter} for the information on reproducibility.
\end{enumerate}

\section{Backgrounds}
\begin{figure*}[h!]
\begin{center}
\subfigure[DLinear]{{\includegraphics[width=0.6\columnwidth]{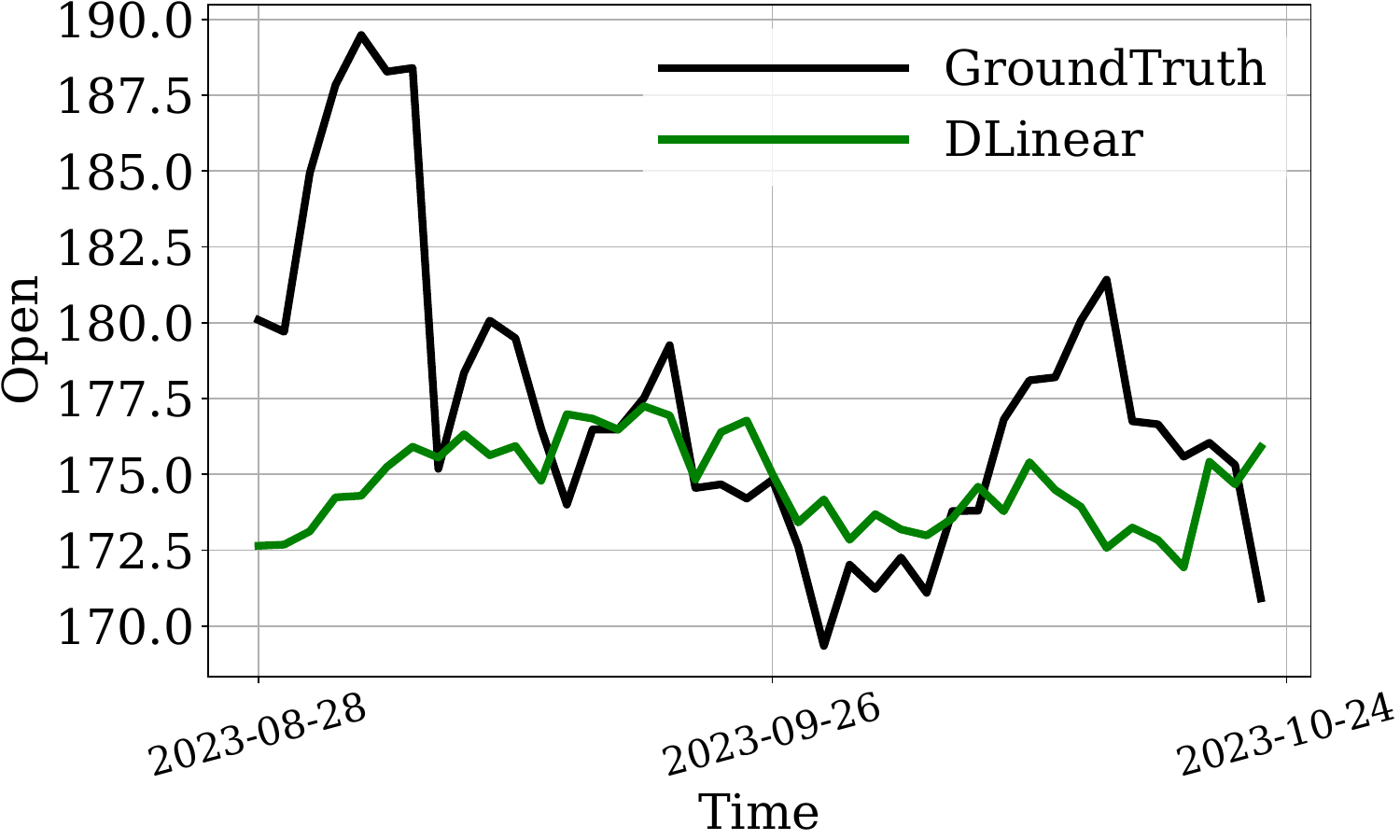}} }\hfill
\subfigure[PatchTST]{{\includegraphics[width=0.6\columnwidth]{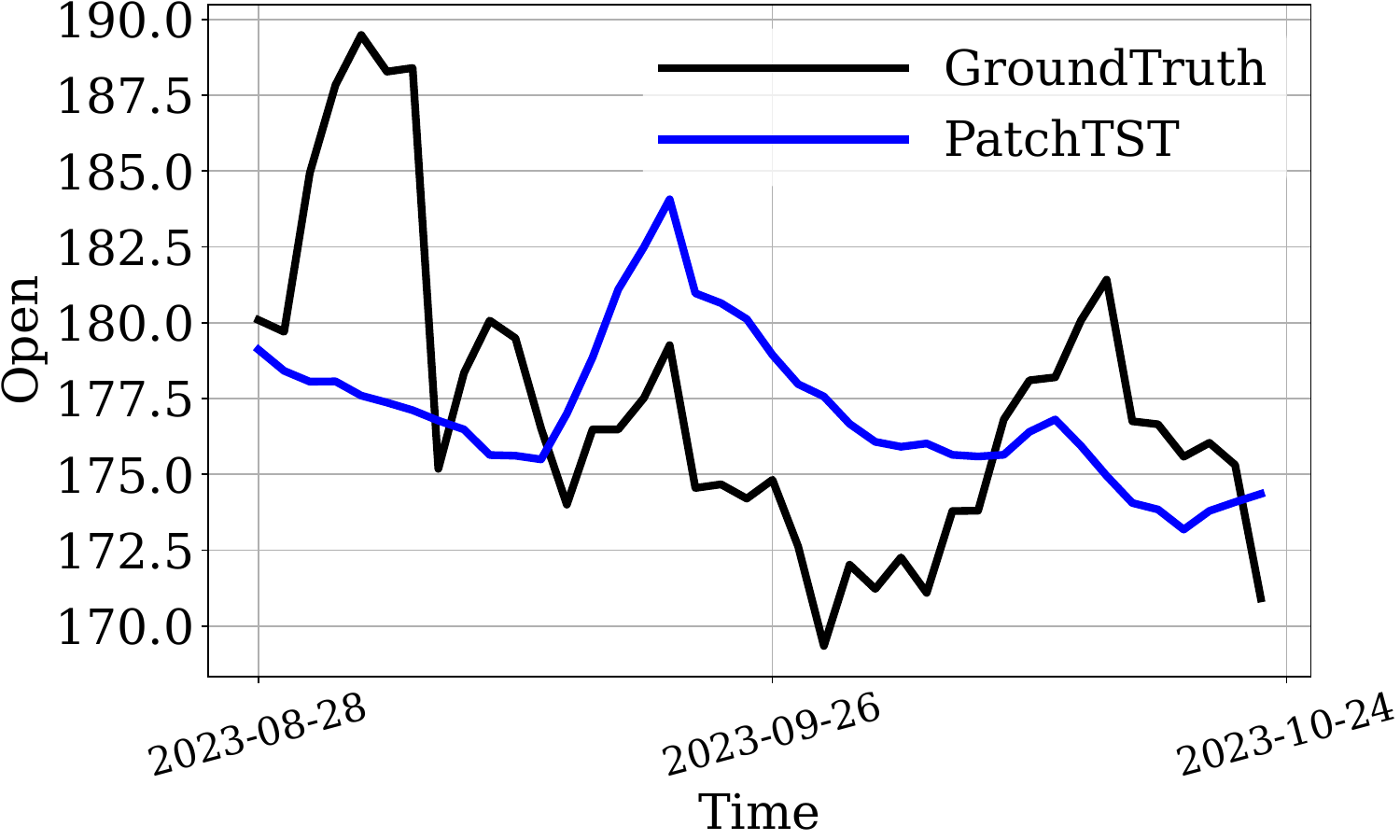}} }\hfill
\subfigure[CONTIME]{{\includegraphics[width=0.6\columnwidth]{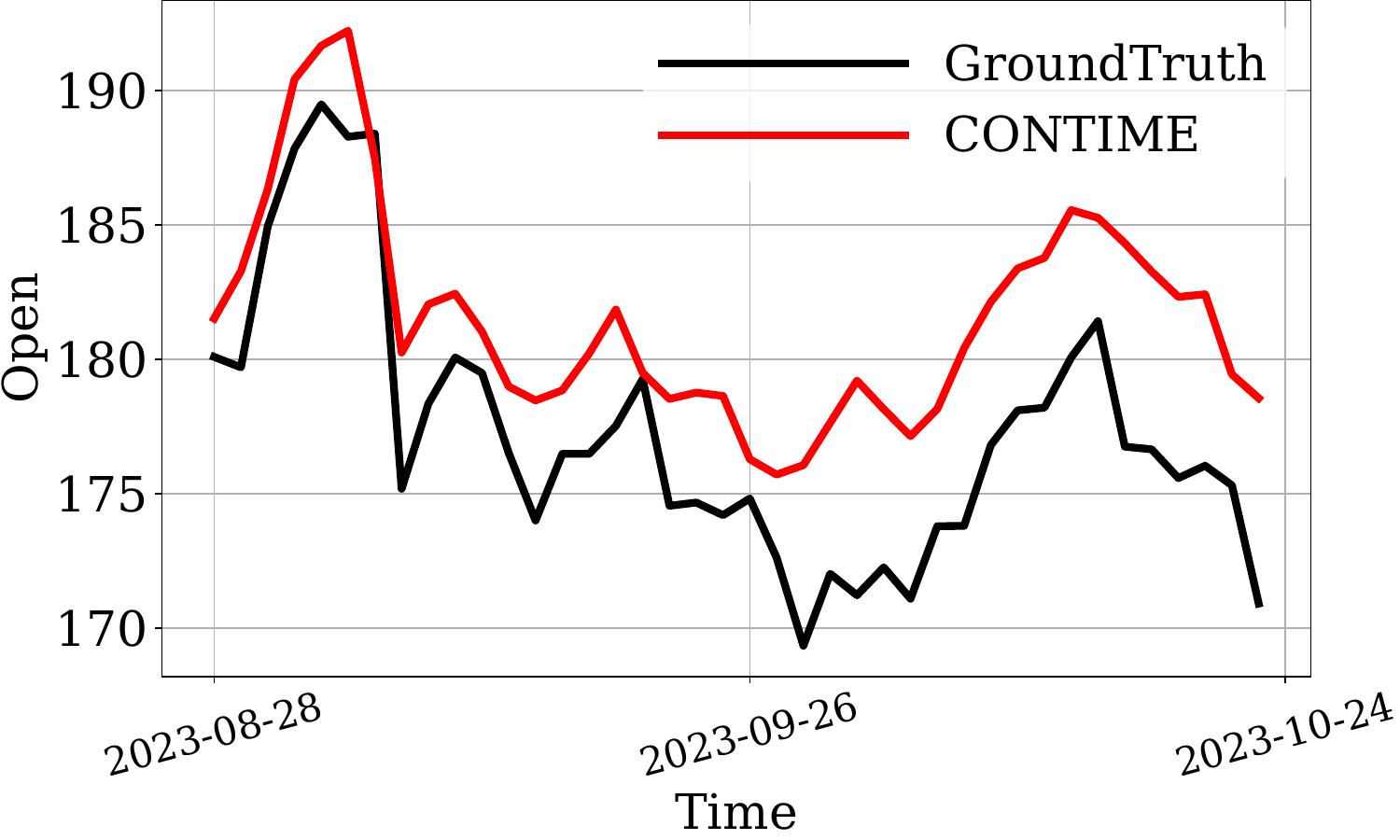}} }\hfill
\caption{Visualization for comparing characteristics of each metric (MSE, DTW, TDI). Forecasting results on AAPL from August 28th, 2023 to October 24th, 2023.}
\label{fig:teaser_metric}
\end{center}
\end{figure*}
\subsection{Time series forecasting models}
In this section, we introduce various time series forecasting models from ODE-based models to recent models.

\paragraph{ODE-based Models:} Neural ODE enable the processing of time-series data in a \emph{continuous} manner, allowing them to read and write values at any arbitrary time-point $t$ through the differential equation presented in Equation~\eqref{eq:node}.

\begin{align}\label{eq:node}
    \mathbf{h}(T) = \mathbf{h}(0) + \int_{0}^{T} f(\mathbf{h}(t), t;\mathbf{\theta}_f) dt,
\end{align}where $\mathbf{h}(t) \in \mathbb{R}^D$, $t \in [0,T]$, represents a $D$-dimensional vector (with boldface denoting vectors and matrices). The derivative $\dot{\mathbf{h}}(t) \stackrel{\text{def}}{=} \frac{d\mathbf{h}(t)}{dt}$ is approximated by the neural network $f(\mathbf{h}(t), t;\mathbf{\theta}_f)$, and solving the initial value problem yields the final value $\mathbf{h}(T)$ from the initial value $\mathbf{h}(0)$. The ODE-based neural network learns by estimating the differential values of the data function $f(\mathbf{h}(t), t;\mathbf{\theta}_f)$ using ODE solvers such as the explicit Euler method, the 4th order Runge-Kutta (RK4) method, the Dormand-Prince (DOPRI) technique, and similar approaches~\citep{chen2018neural}.

There also exist prominent time-series processing models based on NODE, such as Neural Controlled Differential Equation (NCDE). NCDE, an advanced network of NODE, utilizes the Riemann--Stieltjes integral, as shown in Equation~\eqref{eq:ncde}. Unlike NODE, which employs the Riemann integral, NCDE can continuously read $X(t)$ values over time. Thus, NCDE can overcome the limitations of NODE that depend on initial values~\citep{kidger2020neural}. 
\begin{align}
    \begin{split} 
    \label{eq:ncde}
    \mathbf{h}(T) &= \mathbf{h}(0) + \int_{0}^{T} f(\mathbf{h}(t);\mathbf{\theta}_f) dX(t)\\
    &= \mathbf{h}(0) + \int_{0}^{T} f(\mathbf{h}(t);\mathbf{\theta}_f) \frac{dX(t)}{dt} dt,
    \end{split}
\end{align}
Moreover, NCDE creates a continuous path $X(t)$ by employing interpolation techniques like the natural cubic spline or Hermite cubic spline. Since the natural cubic spline uses all time observations to form a continuous path $X(t)$, so in the context of time-series forecasting, the natural cubic spline method may not be suitable for forecasting tasks. Therefore, in this paper, we opt for the Hermite cubic spline method to generate the continuous path $X(t)$~\citep{morrill2021neural}.

\paragraph{Transformer-based models:}

Subsequent advancements have introduced transformer architectures, originally devised for natural language processing\citep{vaswani2023attention}, to the domain of time series forecasting, thereby incorporating self-attention mechanisms. These models, utilizing self-attention, have demonstrated remarkable efficacy in capturing overarching dependencies within sequential data, leading to the development of significant transformer-based studies such as Autoformer and FEDformer~\citep{wu2021autoformer, zhou2022fedformer}. While Autoformer employs auto-correlation attention for periodic patterns, it falls short in series decomposition, overly depending on a basic moving average for detrending, which may constrain its ability to capture intricate trend patterns. On the other hand, FEDformer~\cite{zhou2022fedformer} integrates the Transformer with seasonal trend decomposition, utilizing decomposition for global profiles and Transformers for detailed structures. Despite these notable accomplishments, it is crucial to acknowledge that transformer-based architectures exhibit inefficiencies in capturing local dependencies and temporal information. This constraint has spurred continuous research endeavors aimed at addressing and improving the effectiveness of transformer-based models in comprehensively capturing both global and local intricacies within time series data.

\paragraph{Recent state-of-the-art models:}
The innovative introduction of PatchTST \citep{Yuqietal-2023-PatchTST} represents a groundbreaking approach that employs patch-based representations to enhance the capture of both local and global patterns within time series data. Building upon this, PatchTST further enhances its methodology by segmenting time series before utilizing a Transformer, demonstrating superior performance compared to existing models. Despite being rooted in the foundational Transformer architecture, innovations are focused on transitioning from self-attention to sparse self-attention, often overlooking a comprehensive global view of time series data. DLinear~\citep{zeng2023transformers} has significantly contributed to the field by exploring linear models for time series forecasting. In defiance of the prevalent assumption that only highly complex nonlinear models excel in this context, DLinear has exhibited competitive performance with a linear layer, emphasizing efficiency and interoperability.



In summary, the progression from Neural ODE to PatchTST and DLinear signifies an ongoing quest for more effective and efficient deep learning models in the domain of time series forecasting. Each model brings unique features, methodologies, and challenges that challenge prevailing assumptions, with a notable emphasis on a novel approach for model evaluation based on the MSE.

\subsection{Evaluation and training metrics}

In the realm of evaluating and training deep models for time series forecasting, prevalent approaches heavily depend on metrics such as MAE, MSE, and their variants, including SMAPE. While these metrics effectively gauge overall model performance, evaluating shape and temporal location is crucial for a more comprehensive assessment. Techniques like Dynamic Time Warping (DTW)~\citep{1163055} are employed to capture shape-related metrics, and Temporal Distortion Index (TDI)~\citep{vallance2017towards,frias2017assessing} is utilized for prediction delay estimation. However, due to their non-differentiability, these evaluation metrics are unsuitable as loss functions for training deep neural networks.

Addressing the challenge of optimizing non-differentiable evaluation metrics directly, the development of surrogate losses has been explored across various domains, including computer vision. Recently, alternatives to MSE have been investigated, with a focus on seamless approximations of DTW~\citep{cuturi2017soft} to train deep neural networks. Despite its effectiveness in assessing shape errors, the inherent design of DTW, i.e., the invariance to elastic distortions, overlooks crucial considerations about the temporal localization of changes. \citet{le2019shape} attempted to train models with a loss that combines DTW and TDI to account for both the shape and temporal distortion.

\begin{table}[ht!]
\small \centering
\caption{Effect of Metrics on Figure \ref{fig:teaser_metric}. \textit{Time} and \textit{Shape} respectively denote the resemblance in timing and shape between the ground-truth and the prediction. (O/X means whether each result displays good output when scrutinized with visualization)}
\label{tab:teaser_2}
\begin{tabular}{ccccccc} \toprule
\multirow{2}{*}{Models} & \multicolumn{2}{c}{\textit{Time}} & \multicolumn{2}{c}{\textit{Shape}} & \multicolumn{2}{c}{Score}\\   \cmidrule(lr){2-3} \cmidrule(lr){4-5} \cmidrule(lr){6-7}
\multicolumn{1}{c}{} & \multicolumn{1}{c}{TDI} & \multicolumn{1}{c}{Vis} & \multicolumn{1}{c}{DTW} & \multicolumn{1}{c}{Vis} & \multicolumn{1}{c}{MSE} & \multicolumn{1}{c}{Vis} \\ \toprule
DLinear   & 3.810 & X & 1.409 & X & 0.084 & O \\ 
PatchTST  & 3.166 & X & 1.253 & O & 0.091 & O                       \\ 
CONTIME   & 2.378 & O & 1.114 & O & 0.074 & O                       \\ \bottomrule
\end{tabular}
\end{table}

Figure \ref{fig:teaser_metric} and Table \ref{tab:teaser_2} provide examples where each metric (MSE, DTW, TDI) excels in analyzing experimental results (refer to Table \ref{tbl:exp_1}). By examining the relationship between metric values and visualization results, we gain insight into the role of each metric.
In Figure \ref{fig:teaser_metric}, the results of DLinear in Figure \ref{fig:teaser_metric}.(a) demonstrate a relatively small MSE of $0.084$, yet exhibit a lack of superiority in terms of prediction shape and timing. This observation is further supported by the TDI and DTW metrics. It illustrates that a good MSE score does not necessarily guarantee accurate time series prediction. Figure \ref{fig:teaser_metric}.(b) presents predictions with a more accurate shape than Figure \ref{fig:teaser_metric}.(a), but entails a prediction delay in determining the direction of movement. Consequently, TDI and MSE values are large compared to smaller DTW values. This indicates that time series forecasts cannot be evaluated solely using DTW and MSE. Figure \ref{fig:teaser_metric}.(c) showcases the prediction results of CONTIME, demonstrating excellent performance in terms of time series shape and timing, naturally leading to small MSE values. These analyses emphasize the necessity to evaluate time series forecasts from diverse perspectives.

This paper advances this approach by directly computing the gradient of sequences. To enable the instantaneous prediction of rises and declines, we incorporate a regularization component that utilizes time-derivatives. This strategy addresses the gap by introducing time-derivative regularization to the traditional MSE loss. By decoupling the training for prediction delay and the MSE criterion, this paper aims to provide a robust framework for training deep neural networks on real-time series data.

\subsection{The prediction delay in time series forecasting}

Time series forecasting is a crucial task spanning diverse domains, including finance and environmental science. A significant challenge in this domain is prediction delay, where models may struggle to provide accurate and timely predictions. This subsection explores existing research addressing prediction delay in time series forecasting, emphasizing neural network approaches and other relevant methodologies. In \citep{conway1998delayed}, challenges in time series forecasting using neural networks are investigated, and strategies to mitigate forecast delays are proposed. The study assesses the impact of delay on prediction accuracy and explores techniques to enhance prediction timeliness using neural network architectures. In addition, three studies in~\citep{dixit2015removing,han2023economic,long2024unveiling} focus on applying artificial neural networks to rainfall-runoff modeling, economic models, traffic forecasting, and so on and investigating constraints related to forecast delays. One of the studies evaluates the trade-off between hydrological state representation and model evaluation, emphasizing the challenges posed by delays in hydrological forecasting. Beyond the applications like rainfall runoff or wave height predictions, the delay phenomenon is also observed in the economic field.

\paragraph{Causes of the prediction delay:} 
We categorize two common causes of prediction delay in time series prediction models:
\begin{enumerate}
    \item Time series data often exhibit temporal dependence, where current values are influenced by past observations. The prediction delay can occur if the model fails to accurately capture these dependencies or experiences a delay in incorporating relevant historical information.
    \item The prediction delay may arise from MSE-based forecasting models' limited ability to adjust to sudden changes in the time series because their primary objective is to minimize the mean square difference between predicted and actual values~\citep{le2019shape}.
\end{enumerate}

In this paper, we propose CONTIME, a model architecture for supervising time-derivatives to eliminate prediction delays. We add a time-derivative regularization to the main task-dependent loss function to effectively handle prediction delays.


\section{Proposed Method}
In this section, we describe our proposed method to address the prediction delay that is common in time series forecasting. In other words, the model we propose is a NODE-based bi-directional continuous GRU that can be explicitly supervised for time-derivatives to address the prediction delay.
\begin{figure}[t]
\vskip 0.2in
\begin{center}
{\includegraphics[width=0.7\columnwidth]{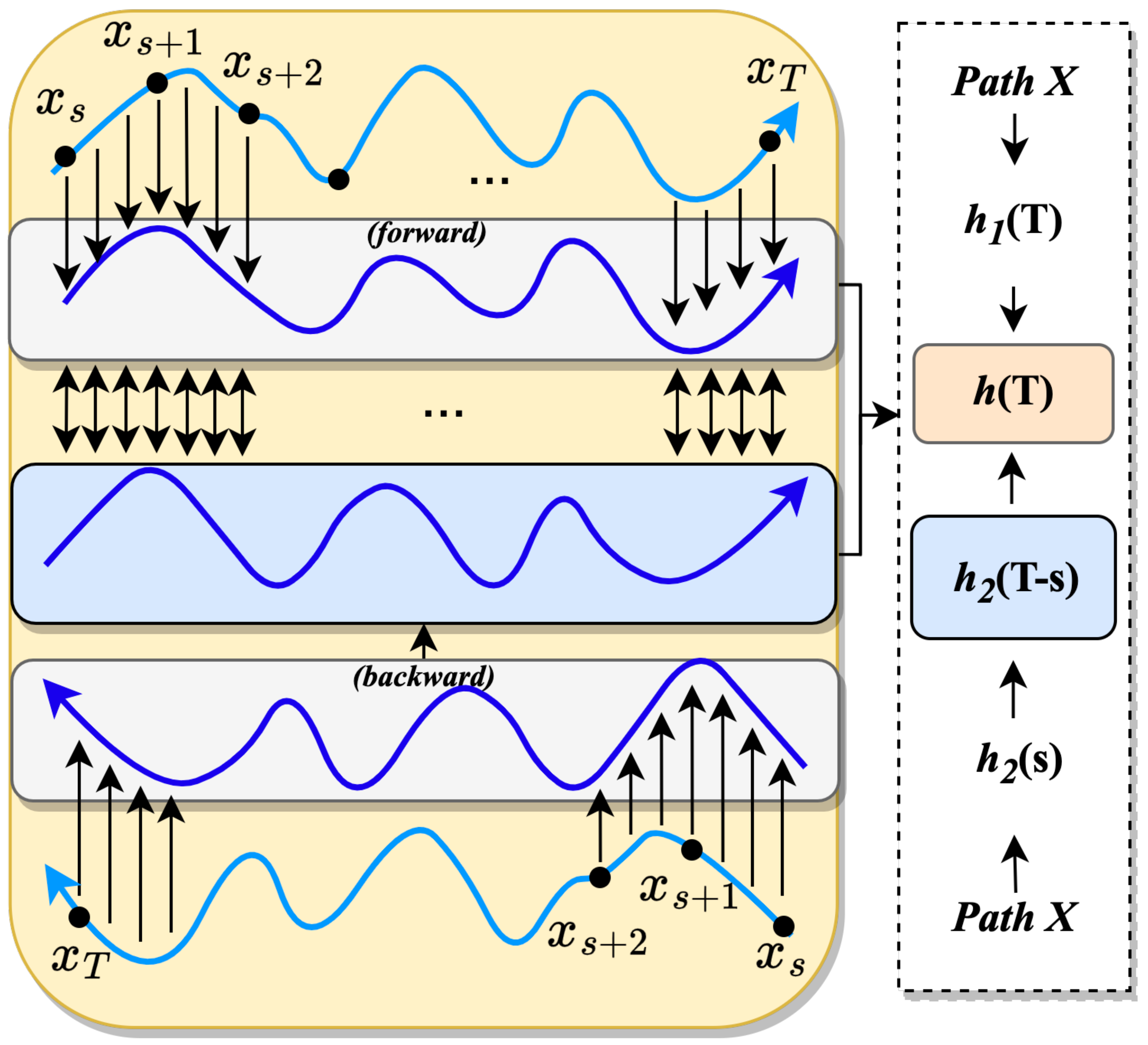}} 
\caption{Overall Architecture}
\label{fig:overall_architecture}
\end{center}
\end{figure}
\subsection{Overall workflow} 
Figure~\ref{fig:overall_architecture} shows the detailed design of our method, CONTIME. The overall workflow is as follows:
\begin{enumerate}
    \item The path $X$ in Figure~\ref{fig:overall_architecture} is created from a discrete time-series sample by Hermite-cubic-spline algorithm. 
    \item CONTIME has a bi-directional structure. As shown in Figure~\ref{fig:overall_architecture}, we perform bi-directional integral operations (highlighted in the gray box in Figure~\ref{fig:overall_architecture}) are conducted both forward ($s \rightarrow T$) and backward ($T \rightarrow s$).
    \item After the forward and backward operations are performed, the hidden vector $\mathbf{h}_2(s)$ from the backward operation is rearranged in the forward direction ($\mathbf{h}_2(T-s)$) through the reverse layer (highlighted in the blue box in Figure~\ref{fig:overall_architecture}).
    \item We can get final hidden vector $\mathbf{h}(T)$ by adding $\mathbf{h}_1(T)$ and $\mathbf{h}_2(T-s)$.
    \item From the hidden vector $\mathbf{h}(T)$, the linear network produces the future predictions.
    \item We explicitly calculate $\frac{d\hat{Y}}{dt}$ from the forecasting prediction $\hat{Y}$ to supervise the time-derivative ($L_{\Delta t}$).
    \item There is a loss to maintain the accuracy of the existing time series predictions ($L_{Task}$) and the time-derivative regulation term ($L_{\Delta t}$) to prevent prediction delay.
\end{enumerate}
We describe each part in detail, followed by a theoretical result that shows training the proposed model is well-posed problem. 
\subsection{Bi-directional CONTIME}\label{sec:bi_d_contime}
We first introduce our formulation to define the proposed CONTIME. The entire module can be written, when we adopt the proposed bi-directional continuous GRU strategy to supervise time-derivative, as follows:
\begin{align}\begin{split} \label{eq:bi_directional_contime}
    \mathbf{h}_1(T) &= \mathbf{h}_1(s) + \int^T_s f_1(\mathbf{h}_{1}(t),t;\mathbf{\theta}_{f_1}) dt, \\  \mathbf{h}_2(s) &= \mathbf{h}_2(T)+ \int^s_T f_2(\mathbf{h}_2(t),t;\mathbf{\theta}_{f_2})dt, \\
\end{split}\end{align} 

where $s$ denotes initial point in time-series sample $X = (X_{s}, ..., X_{T}) \in \mathbb{R}^{(T-s) \times F}$, where $F$ means the number of features. $\mathbf{h}_1(s) = \Phi_{\mathbf{h}_1(X_s)}$, $\mathbf{h}_2(T) = \Phi_{\mathbf{h}_2(X_{T})}$ and 
$\Phi_{\mathbf{h}_1}, \Phi_{\mathbf{h}_2}$ is a fully-connected layer-based feature extractor. In Equation~\eqref{eq:bi_directional_contime}, we use bi-directional integral operations in the forward ($s \rightarrow T$) and backward ($T \rightarrow s$) directions to generate a more useful hidden representation in long sequences. After reverse $\mathbf{h}_2(s)$ to $\mathbf{h}_2(T-s)$, we can write our final hidden representation $\mathbf{h}(T)$ as follows:
\begin{align}
    \begin{split}
        \mathbf{h}(T) &= \mathbf{h}_1(T) + \mathbf{h}_2(T-s).
    \end{split}
\end{align}

In the integration of Equation~\eqref{eq:bi_directional_contime}, we use ODE function $f_1$ and $f_2$ which can be interpreted as $\frac{d\mathbf{h}_1(t)}{dt}$ and $\frac{d\mathbf{h}_2(t)}{dt}$, thereby explicitly calculating the hidden vector $\mathbf{h}(t)$ of GRUs. 

\paragraph{Time-derivative of $\mathbf{h}(t)$:} 
GRUs can be written as follows: 

\begin{align}\begin{split}\label{eq:GRU}
\mathbf{h}(t)&:= \mathbf{z}(t) \odot \mathbf{h}(t-\tau) + (1-\mathbf{z}(t)) \odot \mathbf{g}(t),\\
\mathbf{z}(t) &:= \sigma\big(\mathbf{W}_z X(t) + \mathbf{U}_z \mathbf{h}(t-\tau) + \mathbf{b}_z \big),\\
\mathbf{r}(t) &:= \sigma\big(\mathbf{W}_r X(t) + \mathbf{U}_r \mathbf{h}(t-\tau) + \mathbf{b}_r \big),
\end{split}\end{align} where $\mathbf{W} \in \mathbb{R}^{\dim(\mathbf{h}) \times \dim(\mathbf{x})}$ and $\mathbf{U} \in \mathbb{R}^{\dim(\mathbf{h}) \times \dim(\mathbf{h})}$ are weight matrices, and $\mathbf{b} \in \mathbb{R}^{\dim(\mathbf{h})}$ is a bias vector. $\sigma$ is a sigmoid function and $\phi$ is a hyperbolic tangent function. $\tau > 0$ is a delay factor --- note that $\tau=1$ in the original design of GRUs whereas we interpret it in a continuous manner. Since the hidden state $\mathbf{h}(t)$ is a composite function of $\mathbf{r}(t)$, $\mathbf{z}(t)$, and $\mathbf{g}(t)$, the derivative of $\mathbf{h}(t)$ can be written as follows:

\begin{align}\begin{split}
    \frac{d\mathbf{h}(t)}{dt} &= \frac{d\mathbf{z}(t)}{dt} \odot \mathbf{h}(t-\tau) + \mathbf{z}(t)\odot\frac{d\mathbf{h}(t-\tau)}{dt} \\ &-\frac{d\mathbf{z}(t)}{dt}\odot \mathbf{g}(t) +(1-\mathbf{z}(t))\odot\frac{d\mathbf{g}(t)}{dt},\\
    &= \frac{d\mathbf{z}(t)}{dt} \odot \big(\mathbf{h}(t-\tau) - \mathbf{g}(t)\big) \\ &+ \mathbf{z}(t)\odot\big(\frac{d\mathbf{h}(t-\tau)}{dt} -\frac{d\mathbf{g}(t)}{dt}\big) + \frac{d\mathbf{g}(t)}{dt},\\
&= \frac{d\mathbf{z}(t)}{dt}\odot \mathbf{\zeta}(t,t-\tau) \\
&+ \mathbf{z}(t)\odot \frac{d\mathbf{\zeta}(t,t-\tau)}{dt} + \frac{d\mathbf{g}(t)}{dt},\\
    \end{split}
\end{align}  where $\mathbf{\zeta}(t,t-\tau) = \mathbf{h}(t-\tau)-\mathbf{g}(t)$. 
So, we can write $\frac{d\mathbf{h}(t)}{dt}$ as follows:

\begin{align}\begin{split}
    \frac{d\mathbf{h}(t)}{dt} &= \frac{d(\mathbf{z}(t)\odot \mathbf{\zeta}(t,t-\tau))}{dt}+\frac{d\mathbf{g}(t)}{dt}.
\end{split}
\end{align} \label{eq:derivation_h}

Finally, Equation~\eqref{eq:bi_directional_contime} can be rewritten as follows:
\begin{align}\begin{split}
    \mathbf{h}_1(T) &= \mathbf{h}_1(s) + \int^T_s \frac{d(\mathbf{z}_1(t)\odot \mathbf{\zeta}_1(t,t-\tau))}{dt}+\frac{d\mathbf{g}_1(t)}{dt} dt,\\
    \mathbf{h}_2(s) &= \mathbf{h}_2(T) + \int^s_T \frac{d(\mathbf{z}_2(t)\odot \mathbf{\zeta}_2(t,t-\tau))}{dt}+\frac{d\mathbf{g}_2(t)}{dt} dt.
\end{split}\end{align}
Other derivatives for $\mathbf{z}(t)$, $\mathbf{g}(t)$, and $\mathbf{r}(t)$ are in Appendix~\ref{appendix:derivatives}. 
Finally, the time-derivatives of $\mathbf{h}(t),\mathbf{z}(t),\mathbf{g}(t),$ and $\mathbf{r}(t)$ is written as follows: 

\begin{align} 
&\frac{d}{dt}{\begin{bmatrix}
  \mathbf{h}(t) \\
  \mathbf{z}(t) \\
  \mathbf{g}(t) \\
  \mathbf{r}(t)
  \end{bmatrix}\!}  := {\begin{bmatrix}
  \frac{d(\mathbf{z}(t)\odot \mathbf{\zeta}(t,t-\tau))}{dt}+\frac{d\mathbf{g}(t)}{dt}\\
  \sigma\big(\mathbf{A}(t,t-\tau))(1-\sigma(\mathbf{A}(t,t-\tau))\big) \frac{d\mathbf{A}(t,t-\tau)}{dt} \\
  \big(1-\phi^2(\mathbf{B}(t,t-\tau)\big)\frac{d\mathbf{B}(t,t-\tau)}{dt} \\
  \sigma\big(\mathbf{C}(t,t-\tau))(1-\sigma(\mathbf{C}(t,t-\tau))\big) \frac{d\mathbf{C}(t,t-\tau)}{dt} \\
  \end{bmatrix}.\!} \label{eq:derivation_full}
\end{align}

$\frac{dX(t)}{dt}$ contained by the derivatives of $\mathbf{A}, \mathbf{B},$ and $\mathbf{C}$ can also be calculated since we use an interpolation method to construct continuous path $X(t)$ (see Appendix~\ref{appendix:Interpolation}).

We derive the final prediction result $\hat{Y}$ by passing $\mathbf{h}(T)$ to a single fully-connected layer $\texttt{FC}_{\theta_p}$:

\begin{align}
    \hat{Y} &= \texttt{FC}_{\theta_p}(\mathbf{h}(T)),
\end{align}
where $\hat{Y} := (\hat{Y}_1, ..., \hat{Y}_P) \in \mathbb{R}^{P \times F}$ with the prediction length $P$. 
\subsection{Why Continuous GRU?} \label{main:continuous_gru}
In this section, we outline the rationale behind selecting GRU as the primary network architecture for CONTIME.
\paragraph{GRU-based network:} The hidden representation $\mathbf{h}(t)$ in the GRU (Equation~\ref{eq:GRU}) comprises hidden vectors at time $t$ and $t-\tau$. We propose a GRU-based network, called CONTIME, embracing the benefits of modeling past hidden representations in reducing prediction delay. 
\paragraph{$\mathbf{h}(t)$ to $\frac{d\mathbf{h}(t)}{dt}$:} Due to the GRU Equation~\eqref{eq:GRU} including both time $t$ and time $t-\tau$, $h(t)$ is redefined as the derivative of $h(t)$ w.r.t. time in multiple papers proposing GRU-based networks \citep{debrouwer2019gruodebayes,long2024unveiling}. Similarly, we redefine GRU Equation~\eqref{eq:GRU} as $h(t)$ for time, albeit with $\tau < 1$ compared to the conventional GRU where $\tau=1$.
\paragraph{Discrete to continuous:} Essentially, we introduce a continuous ODE-based GRU as opposed to the discrete GRU. This continuous approach facilitates detailed and continuous modeling between discrete time points, allowing a more comprehensive representation of the value of $h(t+\tau)$ between $h(t)$ and $h(t+1)$.
This sublime design well aligns with our objective of supervising time-derivatives and eliminating prediction delays.

\subsection{How to train}\label{main:how_to_train}

Our proposed model, CONTIME, uses a loss based on MSE and a time-derivative regularization to accurately predict time series and prevent prediction delays. The final loss $L_{CONTime}$ is the sum of $L_{Task}$ and our time-derivative loss $L_{\Delta t}$. 
\begin{align}\begin{split}
    L_{Task} &= \text{MSE}(Y,\hat{Y}),\\
    L_{\Delta t} &= \text{MSE}(Y_{\Delta t},\hat{Y}_{\Delta t}),
\end{split}\end{align}where $Y$ is a ground-truth time series and $\hat{Y}$ is an inferred time series. $\Delta t$ denotes their time-derivatives.

\paragraph{$\Delta t$ Loss:} 
The purpose of the $\Delta t$ loss function is to oversee time differentiation. Essentially, it ensures that accurate time series predictions are achieved without any delay by adjusting the increment or decrement pattern. To solve this problem, we explicitly calculate $\frac{d\hat{Y}}{dt}$ as follows: 
{\small\begin{align}\begin{split}\label{eq:dyhatdt}
        &\hat{Y}_{\Delta t}
            = \frac{d(\texttt{FC}_{\mathbf{\theta}_p}(\mathbf{h}(T)))}{dt} = \frac{d(\mathbf{W}_{\mathbf{\theta}_P}(\mathbf{h}(T))+\mathbf{b}_{\theta_p})}{dt} =  \mathbf{W}_{\mathbf{\theta}_p}\frac{d\mathbf{h}(T)}{dt},
\end{split}
\end{align}}since $\texttt{FC}_{\mathbf{\theta}_p}$ is a fully connected layer, $\texttt{FC}_{\mathbf{\theta}_p}(\mathbf{h}(T))$ can be written as $\mathbf{W}_{\mathbf{\theta}_P}(\mathbf{h}(T))+\mathbf{b}_{\theta_p})$ and $\frac{d\mathbf{h}(T)}{dt}$ is defined by the ODE function $f_1$. So, Equation~\eqref{eq:dyhatdt} can be easily calculated by the automatic differentiation method. Therefore, we use the MSE loss between $\hat{Y}_{\Delta t}$ and $Y_{\Delta t}$ to supervise the time-derivative, where $Y_{\Delta t} := Y_{t_i} - Y_{t_{i-1}}$ --- in other words, we use the difference $Y_{\Delta t}$ to supervise the derivative $\hat{Y}_{\Delta t}$, which is reasonable since we do not know the explicit time-derivative of $Y$.  Our loss function can be summarized as follows:
\begin{align}
    L_{CONTIME} = \alpha L_{Task} + \beta L_{\Delta t},
\end{align} \label{eq:loss_teaser}where $\alpha$ and $\beta$ are the coefficients of the two terms. Finally, we can summarize our training algorithm in Algorithm~\ref{alg:train}.

\begin{algorithm}[t]
\SetAlgoLined
\caption{How to train CONTIME}\label{alg:train}
\KwIn{Training data $D_{train}$, Validating data $D_{val}$, Maximum iteration number $max\_iter$}
Initialize $\mathbf{\theta}_{f_1}$, $\mathbf{\theta}_{f_2}$, and other parameters $\mathbf{\theta}_{others}$ if any, e.g., the parameters of the feature extractor; \\
Create a continuous path $X(t)$\; 

$k \gets 0$;

\While {$k < max\_iter$}{ 
    Train $\mathbf{\theta}_{f_1}$ and $\mathbf{\theta}_{f_2}$ and using $L_{CONTIME}$;\\
    Validate and update the best parameters, $\mathbf{\theta}^*_{f_1}$, $\mathbf{\theta}^*_{f_2}$, and $\mathbf{\theta}^*_{others}$, with $D_{val}$\;
    $k \gets k + 1$;
} 
\Return $\mathbf{\theta}^*_{f_1}$, $\mathbf{\theta}^*_{f_2}$, and $\mathbf{\theta}^*_{others}$;
\end{algorithm}

\paragraph{Well-posedness:} The well-posedness\footnote{A well-posed problem means i) its solution uniquely exists, and ii) its solution continuously changes as input data changes.} of NODE was already proved in~\citep[Theorem 1.3]{lyons2004differential} under the mild condition of the Lipschitz continuity. We show that our CONTIME is also well-posed. Almost all activations, such as ReLU, Leaky ReLU, Tanh, Sigmoid, ArcTan, and Softsign, have a Lipschitz constant of 1. Other common neural network layers, such as dropout, batch normalization, and other pooling methods, have explicit Lipschitz constant values. Therefore, the Lipschitz continuity of $\frac{d \mathbf{h}(t)}{dt}$, $\frac{d \mathbf{r}(t)}{dt}$, $\frac{d \mathbf{z}(t)}{dt}$, and $\frac{d \mathbf{g}(t)}{dt}$ can be fulfilled in our case. Accordingly, it is a well-posed problem. Thus, its training process is stable in practice.

\section{Experiments} 
In this section, we describe our experimental environments and results. We conduct experiments on multivariate time series forecasting. All experiments were conducted in the same software and hardware environments. \textsc{Ubuntu} 18.04 LTS, \textsc{Python} 3.8.0, \textsc{Numpy} 1.22.3, \textsc{Scipy} 1.10.1, \textsc{Matplotlib} 3.6.2, \textsc{PyTorch} 2.0.1, \textsc{CUDA} 11.4, \textsc{NVIDIA} Driver 470.182.03 i9 CPU, and \textsc{NVIDIA RTX A5000}. We repeat training and testing procedures with three different random seeds and report their mean scores. We report standard deviations of all 6 datasets in the arXiv version.
\subsection{Experimental settings}
 We list all the descriptions of datasets and detailed experimental settings in Appendix, \ref{appendix:datasets}, and \ref{appendix:hyperparameter}. 
\paragraph{Baselines:} We test the following state-of-the-art baselines to compare our proposed CONTIME with 6 baseline models. (1) DLinear~\citep{zeng2023transformers} is a simple linear network with time series decomposition method and shows state-of-the-art performance. (2) Neural ODE (NODE~\citep{chen2018neural}) is a continuous-time model that defines the hidden state $\mathbf{h}(t)$ with an initial value problem (IVP). (3) Neural CDE (NCDE~\citep{kidger2020neural}) is a conceptually enhanced model of NODE based on the theory of controlled differential equations. (4) Autoformer~\citep{wu2021autoformer} is a transformer-based method which uses an auto-correlation attention for periodic patterns. (5) FEDformer~\citep{zhou2022fedformer} is a transformer-based method which integrates transformer with seasonal trend decomposition, leveraging decomposition for global profiles and transformers for detailed structures. (6) PatchTST~\citep{Yuqietal-2023-PatchTST} is a time series forecasting technique that makes use of patch-based processing to improve the model's capacity to grasp complex patterns and relationships by segmenting temporal sequences into smaller patches.

\paragraph{Datasets:} We evaluate the performance of the proposed CONTIME on six benchmarked datasets, including weather, exchange, and four Stock datasets (AAPL, AMZN, GOOG, MSFT). Among the benchmarked datasets used, weather and exchange are widely utilized and are publicly available at ~\cite{wu2021autoformer}. The Stock dataset has been actively used in ~\cite{jhin2023attentive}. The following is a description of the six experimental data sets. (1) The \textbf{Stocks} dataset~\cite{appl,aws,goog,msft} contains stock prices of four companies (APPLE, AMAZON, Google, and Microsoft). All four datasets measure 6 stock indicators (Open price, High price, Low price, Close price, Adj Close price, and Volume) of each company from January 17th, 2019 to January 4th, 2024. (2) \textbf{Exchange} contains exchange data among 8 countries~\cite{lai2018modeling}. (3) \textbf{Weather} is data that measures 21 weather indicators, including temperature and humidity, every 10 minutes throughout 2020. 

\begin{table*}[t]
\centering 
\scriptsize
\caption{Experimental results on 6 datasets. The best results are in \textbf{bold} and the second best are \underline{underlined}.}\label{tbl:exp_1}
\begin{tabular}{ccccccccccccccccccccccccc} \toprule
\multicolumn{2}{c}{Datasets}   &  \multicolumn{3}{c}{APPL} & \multicolumn{3}{c}{AMZN} & \multicolumn{3}{c}{GOOG} & \multicolumn{3}{c}{MSFT} & \multicolumn{3}{c}{Exchange} & \multicolumn{3}{c}{Weather}\\\cmidrule(lr){1-2} \cmidrule(lr){3-5} \cmidrule(lr){6-8} \cmidrule(lr){9-11} \cmidrule(lr){12-14} \cmidrule(lr){15-17} \cmidrule(lr){18-20}
 & $P$   & TDI  & DTW  & MSE  & TDI & DTW & MSE  & TDI & DTW & MSE   & TDI & DTW & MSE & TDI & DTW & MSE & TDI & DTW & MSE   \\\midrule
\multirow{4}{*}{\begin{sideways}DLinear\end{sideways}} 
& 24    & 3.180 & 1.409  & \underline{0.084}  
        & 3.855 & 2.239  & 0.265   
        & 3.766 & \underline{1.297}  & \underline{0.166}  
        & 4.327 & \textbf{1.430} & 0.197 
        & 3.629 & \textbf{0.533} & \textbf{0.044}
        & 3.505 & 1.894 & \underline{0.119}\\
& 36    & 5.106 & 1.940 & 0.187
        & \underline{5.396}  & 2.726  & 0.372 
        & \underline{4.835}  & 2.229  & 0.199    
        & 6.103 & 2.385 & 0.319
        & 5.638 & \underline{0.781} & \underline{0.065} 
        & 5.944 & \underline{1.436} & \underline{0.144} \\
& 48    & 7.751  & 2.323  & 0.213  
        & 8.915  & 2.964  & 0.408  
        & \underline{7.518}  & 2.568 & 0.262   
        & \underline{7.324} & 3.738 & 0.468
        & 7.989  & \underline{1.742} & \textbf{0.084}
        & 8.208  & 1.817  & \underline{0.161}\\
& 60    & 10.84  & 2.907  & 0.258  
        & 9.252  & 3.017  & 0.347 
        & 12.39  & 2.848  & 0.294   
        & 12.10  & 4.247  & 0.492 
        & 11.01  & 2.304  & \underline{0.107} 
        & 10.16  & \textbf{1.771} &\textbf{0.174}\\\midrule
\multirow{4}{*}{\begin{sideways}NODE\end{sideways}} 
& 24    & 3.739  & 4.330  & 0.168   
        & 3.063  & 3.275  & 0.397  
        & 3.684  & 6.399  & 1.298  
        & 4.596 & 3.389 & 0.359 
        & 2.085 & 2.855 & 0.525 
        & 2.758 & 4.262 & 0.336\\
& 36    & \underline{4.911}  & 2.916  & 0.328 
        & 5.479 & 4.893 & 0.464  
        & 5.793 & 4.223 & 0.646  
        & 6.769 & 4.329 & 0.496
        & \underline{4.055} & 9.289 & 1.137 
        & \underline{4.314} & 6.193 & 1.261\\
& 48    & 7.482 & 4.203 & 0.535  
        & 7.149 & 6.436 & 0.813  
        & 7.795 & 5.112 & 0.794  
        & 8.868 & 4.656 & 0.504 
        & 6.104 & 6.028 & 1.100
        & 6.827 & 6.294 & 1.261\\
& 60    & \underline{8.702} & 10.25  & 1.149  
        & \underline{8.954} & 6.333  & 1.033  
        & \underline{9.513} & 5.648  & 0.874   
        &\underline{10.72} & 7.963 & 0.618
        & 9.822 & 6.621 & 1.056
        & 10.54 & 7.652 & 1.506\\\midrule
\multirow{4}{*}{\begin{sideways}NCDE\end{sideways}} 
& 24    & 5.039  & 4.555  & 0.227
        & \underline{2.984}  & 5.493  & 0.261  
        & 3.719  & 4.601  & 0.517  
        & 4.842 & 2.809 & 0.445 
        & \underline{1.874} & 3.689 & 0.576
        & \underline{2.489} & 5.609 & 0.854\\
& 36    & 6.651  & 3.199  & 0.462  
        & 5.829  & 4.022  & 0.335 
        & 4.946  & 3.541  & 0.582 
        & 6.687  & 2.902 & 0.628
        & 4.184  & 8.137 & 0.542 
        & 4.661 & 4.059 & 0.799\\
& 48    & \underline{7.303}  & 4.028  & 0.440
        & 7.113  & 5.817 & 0.711 
        & 8.132  & 6.161 & 0.756  
        &9.018 & 4.327 & 0.690 
        & \underline{6.012} & 7.957 & 0.874
        &6.922 & 4.682 & 0.783\\
& 60    & 11.47  & 3.882 & 0.459  
        & 9.041  & 7.936 & 1.352 
        & 10.02  & 5.637 & 0.771 
        & 12.35 & 5.221 & 0.766 
        & \underline{8.105} & 6.516 & 0.604
        & 9.900 & 5.882 & 0.989\\\midrule
\multirow{4}{*}{\begin{sideways}Autoformer\end{sideways}}     
& 24    & \underline{3.085}  & 1.551  & 0.150 
        & 3.576  & \textbf{1.485}  & \underline{0.174} 
        & 3.289  & \textbf{1.239}  & 0.167 
        & \underline{4.222} & 1.690 & 0.246 
        & 3.158 & 1.120 & 0.098
        & 2.586 & 1.938 & 0.327\\
& 36    & 6.561  & 1.882  & 0.171 
        & 5.541  & 2.032  & 0.203 
        & 5.782  & 2.210  & 0.199 
        & \textbf{5.111} & 2.474 & 0.288
        & 4.724 & 1.516 & 0.125
        & 4.662 & 2.393 & 9.349\\
& 48    & 9.814  & 2.307  & 0.170 
        & \underline{6.941}  & \underline{2.388}  & 0.219 
        & 7.606 & 2.943  & 0.289  
        &7.335 & \underline{2.810} & \underline{0.287}
        & 8.245 & 1.760 & 0.129 
        & 6.955 & 2.855 & 0.415 \\
& 60    & 13.82 & 2.651  & \underline{0.188} 
        & 9.414 & \textbf{2.723}  & \underline{0.275} 
        & 10.80 & 3.248  & 0.279  
        & 12.14 & 3.668 & 0.380 
        & 10.53 & \textbf{2.026} & 0.139
        & 9.944 & 2.854 & 0.415\\\midrule
\multirow{4}{*}{\begin{sideways}FEDformer\end{sideways}}  
& 24    & 3.417  & 1.396  & 0.129
        & 3.108  & 1.764  & 0.232 
        & \underline{3.154}  & 1.587  & 0.204
        & 4.335 & 1.754 & 0.243 
        &3.311 & 0.887 & 0.079 
        & 2.872 & \underline{1.506} & 0.215\\
& 36    & 6.335  & 1.826  & 0.149 
        & 5.878  & 2.201  & 0.249  
        & 5.311  & \underline{2.203}  & 0.215 
        & 6.794 & 2.505 & 0.304 
        & 5.638 & 1.079 & 0.085 
        & 5.108 & 1.801 & 0.313\\
& 48    & 12.64  & 1.932 & 0.135 
        & 7.664  & 2.691 & 0.289  
        & 8.489  & \underline{2.312}  & 0.225
        & 8.203  & 2.891 & 0.308 
        &7.952 & 1.692 & 0.108 
        &\underline{6.342} & 2.053 & 0.226\\
& 60    & 16.39  & 2.642  & 0.204  
        & 12.84  & 2.980  & 0.354  
        & 12.13  & 2.785  & \underline{0.244}
        &12.76 & \textbf{3.209} & \underline{0.321}
        & 10.68 & 2.714 & 0.128 
        & \underline{9.495} &  2.083 & \underline{0.199}\\\midrule
\multirow{4}{*}{\begin{sideways}PatchTST\end{sideways}}  
& 24    & 3.166  & \underline{1.253}  & 0.091 
        & 3.969  & 1.574  & 0.177 
        & 3.706  & 1.554  & \textbf{0.165} 
        & 4.222  & 1.529  & 0.215 
        & 3.658 & 0.903 & 0.056
        & 3.089 & 1.796 & \underline{0.119}\\
& 36    & 5.358  & \textbf{1.417}  & \underline{0.118} 
        & 6.679  & \textbf{1.733}  & \textbf{0.168} 
        & 4.882  & 2.766  & \underline{0.191}  
        & 6.388 & \textbf{2.154} & \textbf{0.234}
        & 5.603 & \textbf{0.766} & 0.078
        &4.849 & 1.128 & 0.149\\
& 48    & 7.984  & \textbf{1.809}  & \underline{0.130}  
        & 8.706  & 2.521  & \underline{0.220} 
        & 7.840  & 2.342  & \underline{0.203} 
        & 10.49 & 3.075 &0.356 
        & 8.083 & 1.701 & 0.099 
        & 6.687 & \textbf{1.473} & 0.181\\
& 60    & 11.00  & \underline{2.626} & 0.202  
        & 12.24  & 3.475 & \underline{0.275} 
        & 10.64  & \textbf{2.673}  & \underline{0.244} 
        & 14.09 & 3.883 & 0.693
        & 11.32 & 2.210 & \textbf{0.106}
        & 10.40 & \underline{1.988} & 0.229\\\bottomrule
\multirow{4}{*}{\begin{sideways}CONTIME\end{sideways}} 
& 24    & \textbf{2.378}  & \textbf{1.114}  & \textbf{0.074} 
        & \textbf{2.866}  & \underline{1.529}  & \textbf{0.167}  
        & \textbf{3.052}  & 1.541  & \textbf{0.165} 
        & \textbf{4.218} & \underline{1.528} & \textbf{0.184}
        & \textbf{1.761} & \underline{0.884} & \underline{0.049}
        & \textbf{2.254} & \textbf{1.023} & \textbf{0.117}\\
& 36    & \textbf{4.807}  & \underline{1.541}  & \textbf{0.089} 
        & \textbf{5.275}  & \underline{1.881}  & \underline{0.193} 
        & \textbf{4.712}  & \textbf{2.189}  & \textbf{0.189} 
        & \underline{5.371} & \underline{2.334} & \underline{0.256}
        & \textbf{3.488} & 1.221 & \textbf{0.063}
        & \textbf{4.120} & \textbf{1.390} & \textbf{0.136}\\
& 48    & \textbf{7.300}  & \underline{1.912}  & \textbf{0.114} 
        & \textbf{6.844}  & \textbf{2.300}  & \textbf{0.209}  
        & \textbf{7.364}  & \textbf{2.297}  & \textbf{0.188}  
        & \textbf{7.296} & \textbf{2.755} & \textbf{0.262}
        & \textbf{5.366} & \textbf{1.683} & \underline{0.097}
        & \textbf{6.226} & \underline{1.805} & \textbf{0.159}\\
& 60    & \textbf{7.932}  & \textbf{2.625}  & \textbf{0.147}  
        & \textbf{8.885}  & \underline{2.873}  & \textbf{0.239}     
        & \textbf{9.271}  & \underline{2.741}  & \textbf{0.210}  
        & \textbf{11.83} & \underline{3.261} & \textbf{0.292}
        & \textbf{7.452} & \underline{2.139} & 0.125
        & \textbf{9.366} & 2.121 & \textbf{0.174}\\\bottomrule
\end{tabular}
\end{table*}

\paragraph{Evaluation metrics:}
As time-series forecasting is a complicated task, evaluating the prediction result only with MSE or MAE is insufficient. Thus, we include DTW and TDI as additional metrics, which can be interpreted as follows, to analyze the time-series forecasting task from multiple perspectives:
\begin{enumerate}
    \item \textbf{DTW:} We evaluate the difference of the overall shape between $Y$ and $\hat{Y}$ via DTW. In particular, the more volatile the data is, the more emphasis is placed on using these metrics. Small DTW values mean the overall shapes of $Y$ and $\hat{Y}$. However, one pitfall of DTW is that it ignores delays.
    \item \textbf{TDI:} TDI quantifies the disparity between the optimal paths of $Y$ and $\hat{Y}$. Further details on TDI can be found in Appendix~\ref{appendix:TDI}. Utilizing TDI, a metric for assessing temporal distortion, is critical for precise predictions. Smaller TDI values indicate minimal prediction delays, aligning with the objectives of this paper.
\end{enumerate}

\subsection{Experimental results}
In this subsection, we analyze the experimental results of six datasets by dividing them into a total of three evaluation metrics (MSE, DTW, and TDI)~\citep{le2019shape}. Table~\ref{tbl:exp_1} introduces our experimental results for time-series forecasting with 6 datasets from various fields. We also report our time complexity and model usage in Appendix~\ref{appendix:time_model}.


\paragraph{Stocks:} The past five years of stock data for AAPL, AMZN, GOOG, and MSFT exhibit both sharp rises and sharp falls, rendering them suitable for assessing accurate time series predictions across various aspects. PatchTST demonstrates specialization in MSE and surpasses other models based on differential equations and transformers. Autoformer exhibits reasonable DTW scores. Among the differential equation-based models, including CONTIME, the lowest TDI scores are observed. Notably, most baseline models specialize in a single metric, such as MSE or DTW, whereas CONTIME outperforms across all metrics with the lowest standard deviation.


\paragraph{Exchange:} Table~\ref{tbl:exp_1} presents the experimental findings for the Exchange dataset. Most models exhibit small MSE values; conversely, the NODE and NCDE models demonstrate superior TDI performance compared to others, indicating the efficacy of models based on differential equations in addressing prediction delays. Significantly, our suggested model, CONTIME, performs second-best with an MSE difference of only about $0.005$ when compared to DLinear. In addition, CONTIME performs fairly well in TDI, indicating its capacity to efficiently reduce prediction delays.

\paragraph{Weather:} In Table~\ref{tbl:exp_1}, our proposed model demonstrates superiority over all other models across all metrics. While DLinear and PatchTST exhibit reasonable performance in MSE and DTW, CONTIME consistently outperforms them. Specifically, CONTIME shows an average decrease of $0.168$ in TDI compared to the second-best model across all prediction lengths. Furthermore, NODE and NCDE exhibit remarkable performance, reaffirming the efficiency of differential-equation-based models in mitigating prediction delays. Unlike baselines that excel in only one of the three metrics, CONTIME clearly demonstrates the best performance across all.
\begin{figure*}[h!]
\begin{center}
\subfigure[AAPL from June 30th,2023 to November 21th,2023]{{\includegraphics[width=1.0\columnwidth]{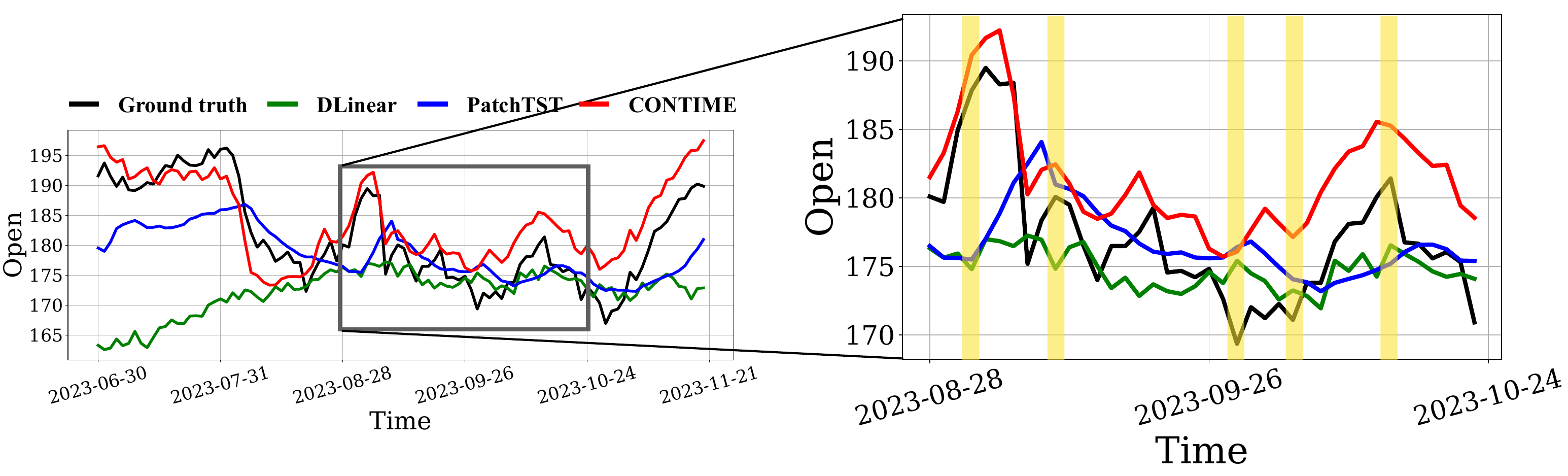}} }\hfill
\subfigure[AMZN from June 30th,2023 to November 21th,2023]{{\includegraphics[width=1.0\columnwidth]{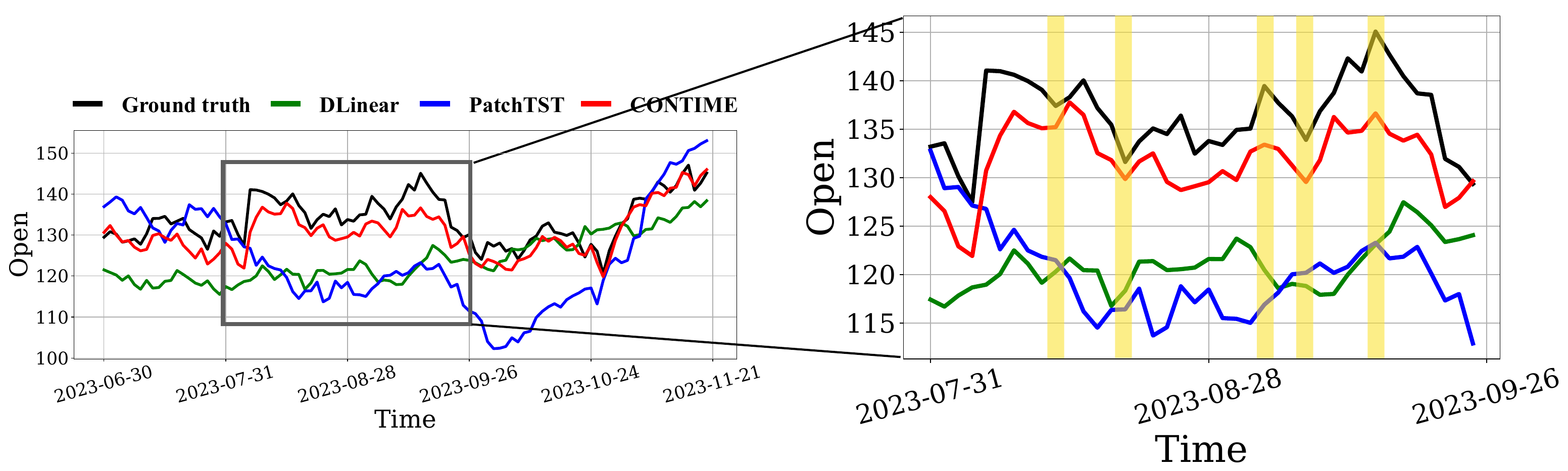}}}\\
\subfigure[Exchange from December 27th,2009 to Feburary 15th, 2010]{{\includegraphics[width=1.0\columnwidth]{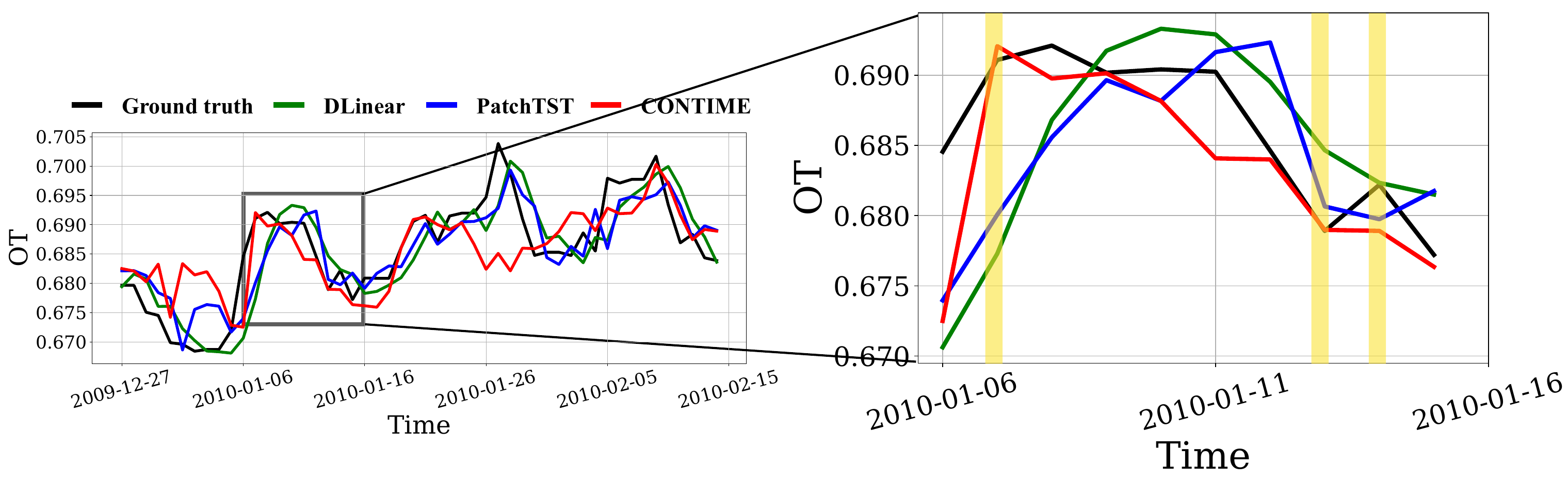}}}\hfill
\subfigure[Weather from 6 hours at December 30th, 2020]{{\includegraphics[width=1.0\columnwidth]{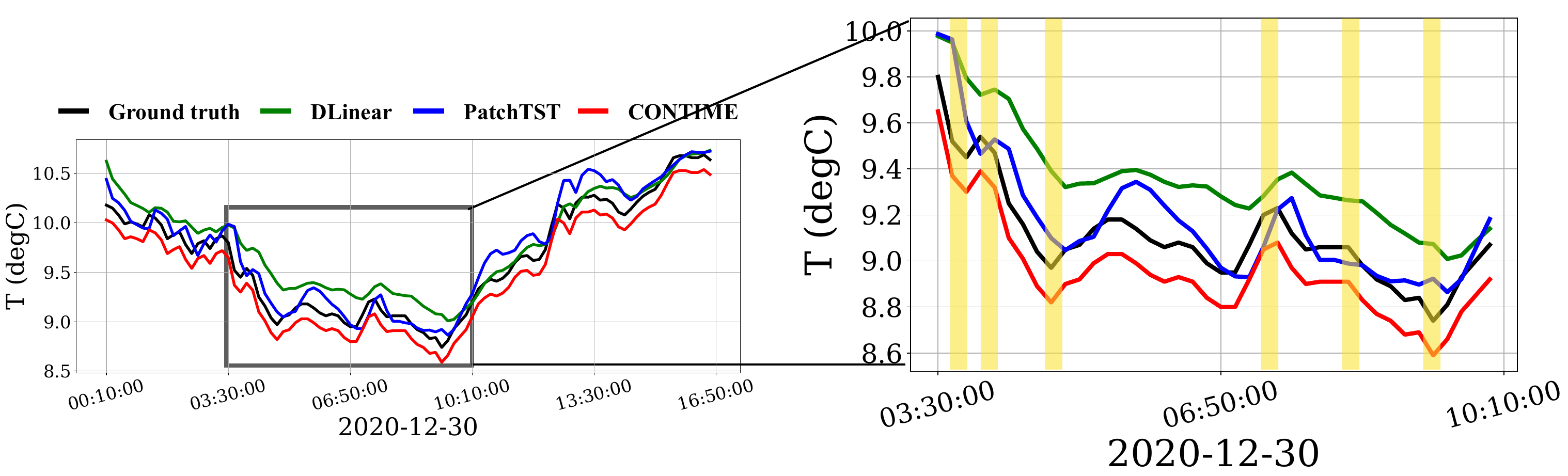}}}
\caption{Forecasting visualization on 4 datasets. More figures are in Appendix~\ref{appendix:forecasting_visualization}}
\label{fig:exp_vis}
\end{center}
\end{figure*}

\subsection{Visualization}
Figure~\ref{fig:exp_vis} provides a visualization of AAPL, AMZN, Exchange, and Weather forecasting results that prove CONTIME's outstanding performance over various aspects compared to state-of-the-art (SOTA) models, such as PatchTST and DLinear. For instance, focusing on the highlighted section in yellow in Figure~\ref{fig:exp_vis}.(a), the SOTA models (blue and green line) predict the opposite of the stock price fluctuation due to delays in the prediction. Specifically, unlike the ground truth, which starts to decline around August, 28th, 2023, state-of-the-art (SOTA) models fail to promptly recognize this change due to a delay. In contrast, CONTIME (red line) accurately captures the actual stock price (black line) in terms of shape, and timing. Figure~\ref{fig:exp_vis}.(b) illustrates the visualization results for the AMZN stock. Across the entire time, CONTIME closely matches the shape of the ground truth and makes predictions without any noticeable delays. Similarly, in Figure~\ref{fig:exp_vis}.(c), while the SOTA models exhibit similarities in terms of shape, their results are delayed; compared to the ground-truth with a high OT value on January 6, 2010, SOTA models predict a high OT value on January 11 due to a delay while CONTIME predicts on time. In Figure~\ref{fig:exp_vis}(d), most models exhibit a shape similar to the ground-truth. Notably, in the highlighted sections of our model (indicated in yellow), the fluctuations of T (degC) are predicted in detail. Conversely, the baseline models made predictions with a slight delay.

\begin{table}[t]
\centering 
\footnotesize
\caption{Comparison on TDI loss and $\Delta t$ loss}\label{tbl:abl_2}
\begin{tabular}{cccccccccccccccc} \toprule
\multirow{2}{*}{Models} & \multirow{2}{*}{$P$}  & \multicolumn{3}{c}{AMZN}  & \multicolumn{3}{c}{Exchange}  \\  \cmidrule(lr){3-5} \cmidrule(lr){6-8}
                         &  & TDI      & DTW      & MSE     & TDI      & DTW      & MSE     \\\midrule 
\multirow{4}{*}{\begin{tabular}[c]{@{}c@{}}CONTIME\\ (Only $L_{Task}$)\end{tabular}}
& 24   & 3.481 & \underline{2.122} & \underline{0.202} & 1.887 & 1.343 & \underline{0.105}\\
& 36   & 5.854 & \underline{1.947} & \underline{0.194} & 3.825 & 1.970 & \underline{0.113}\\
& 48   & 7.775 & \underline{2.255} & \underline{0.214} & 6.301 & 2.836 & \underline{0.242} \\
& 60   & 11.99 & \underline{2.695} & \underline{0.221} & 8.716 & 1.859 & \textbf{0.102}  \\ \midrule
\multirow{4}{*}{\begin{tabular}[c]{@{}c@{}}CONTIME\\ ($L_{Task} + L_{TDI}$)\end{tabular}}
& 24   & \textbf{2.816} & 3.263 & 1.269 & \underline{1.782} & 3.264 & 0.325\\
& 36   & \textbf{4.715} & 2.936 & 0.355 & \underline{3.557} & 4.992 & 0.843\\
& 48   & \underline{6.923} & 4.341 & 0.535 & \textbf{5.229} & 3.596 & 0.852 \\
& 60   & \underline{8.924} & 5.456 & 0.826 & \underline{7.522} & 2.857 & 0.257 \\ \midrule
\multirow{4}{*}{\begin{tabular}[c]{@{}c@{}}CONTIME\\ ($L_{Task} + L_{\Delta t}$)\end{tabular}} 
& 24   & \underline{2.866} & \textbf{1.529} & \textbf{0.167} & \textbf{1.761} & \textbf{0.884} & \textbf{0.049}          \\
& 36   & \underline{5.275} & \textbf{1.881} & \textbf{0.193} & \textbf{3.488} & \textbf{1.221} & \textbf{0.063}          \\
& 48   & \textbf{6.844} & \textbf{2.300} & \textbf{0.209} & \underline{5.366} & \textbf{1.683} & \textbf{0.097}          \\
& 60   & \textbf{8.885} & \textbf{2.873} & \textbf{0.239} & \textbf{7.452} & \textbf{2.139} & \underline{0.125}         \\\bottomrule
\end{tabular}
\end{table}

\begin{figure*}[t]
\vskip 0.1in
\begin{center}
{\includegraphics[width=0.57\columnwidth]{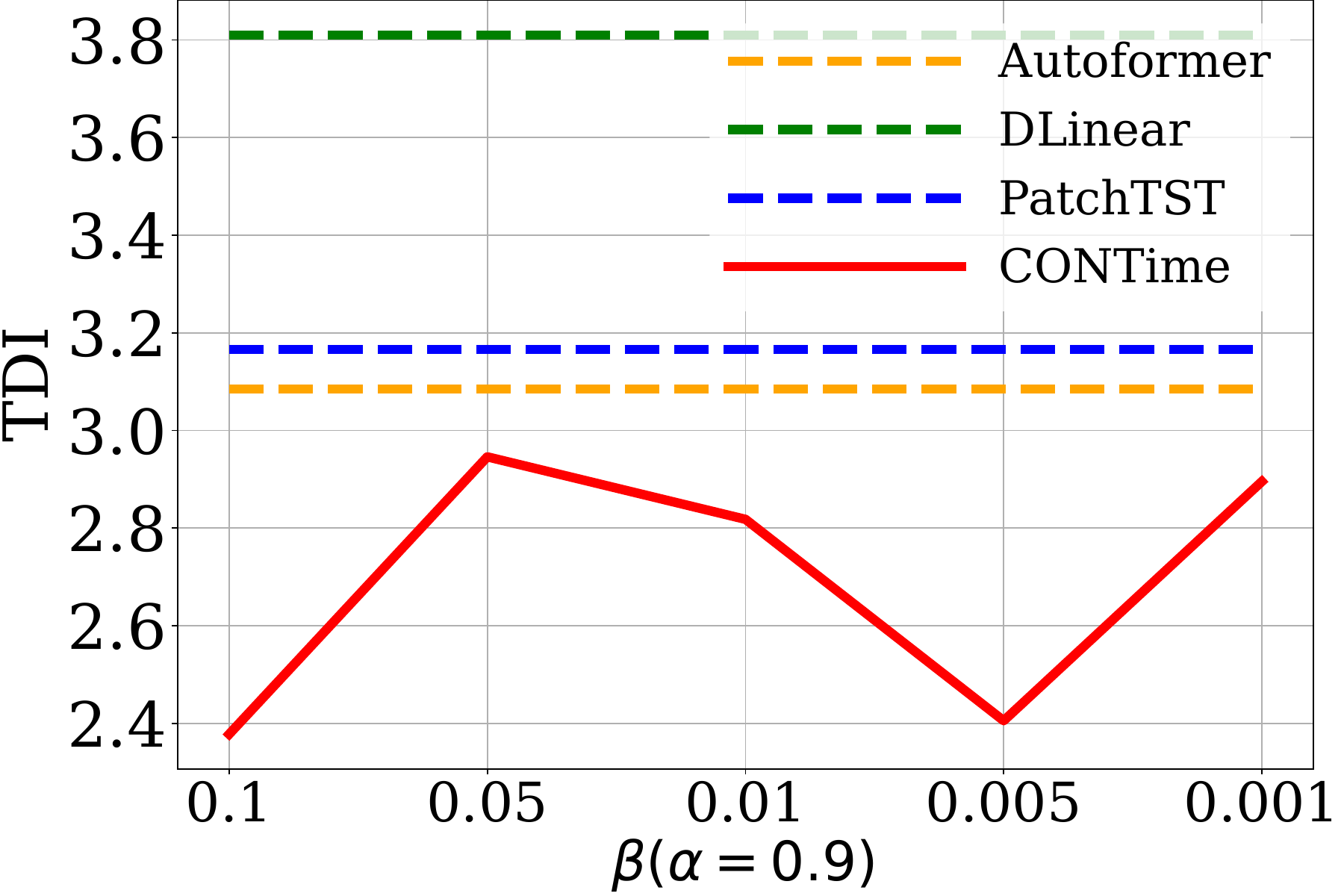}} \hfill
{\includegraphics[width=0.57\columnwidth]{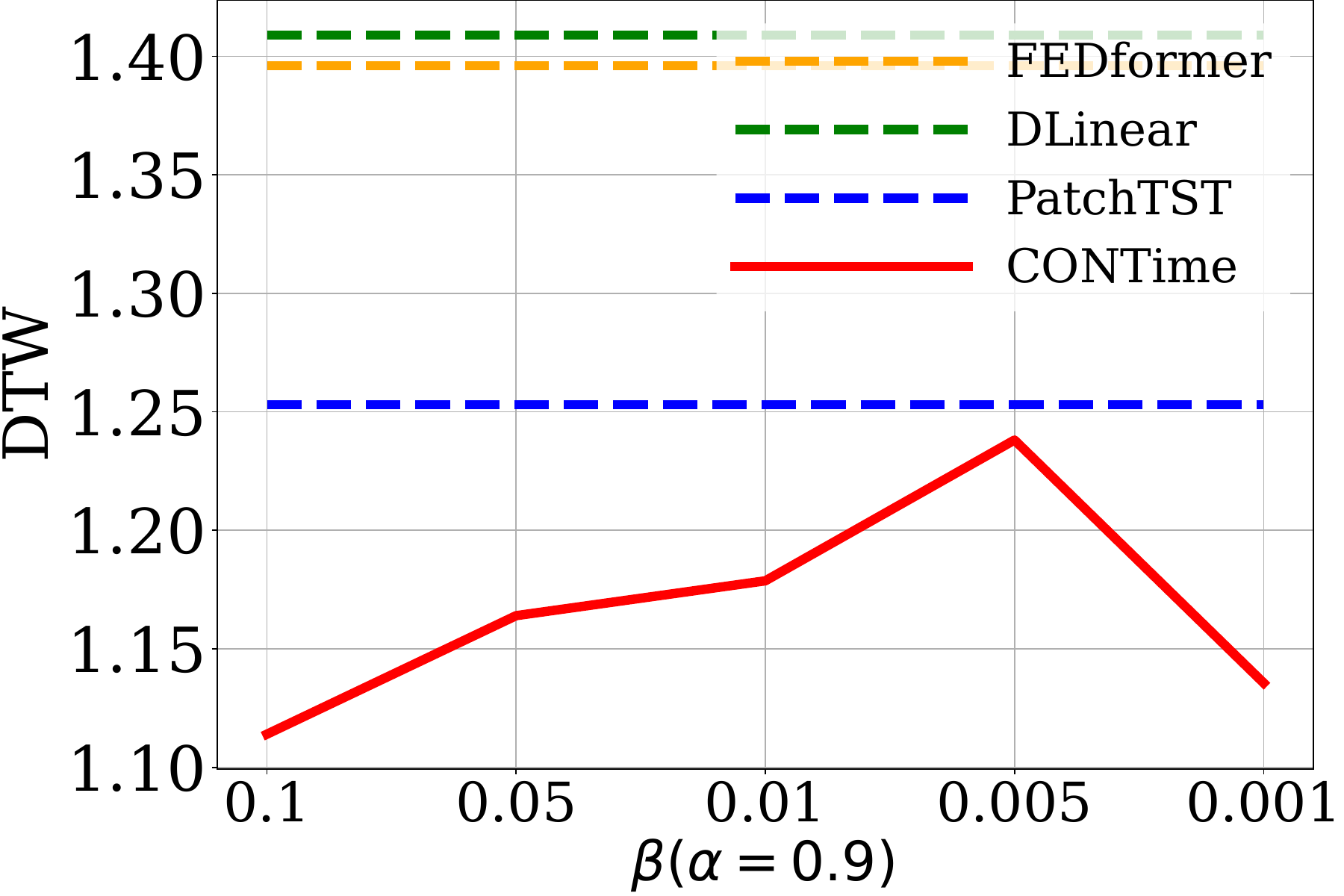}} \hfill
{\includegraphics[width=0.57\columnwidth]{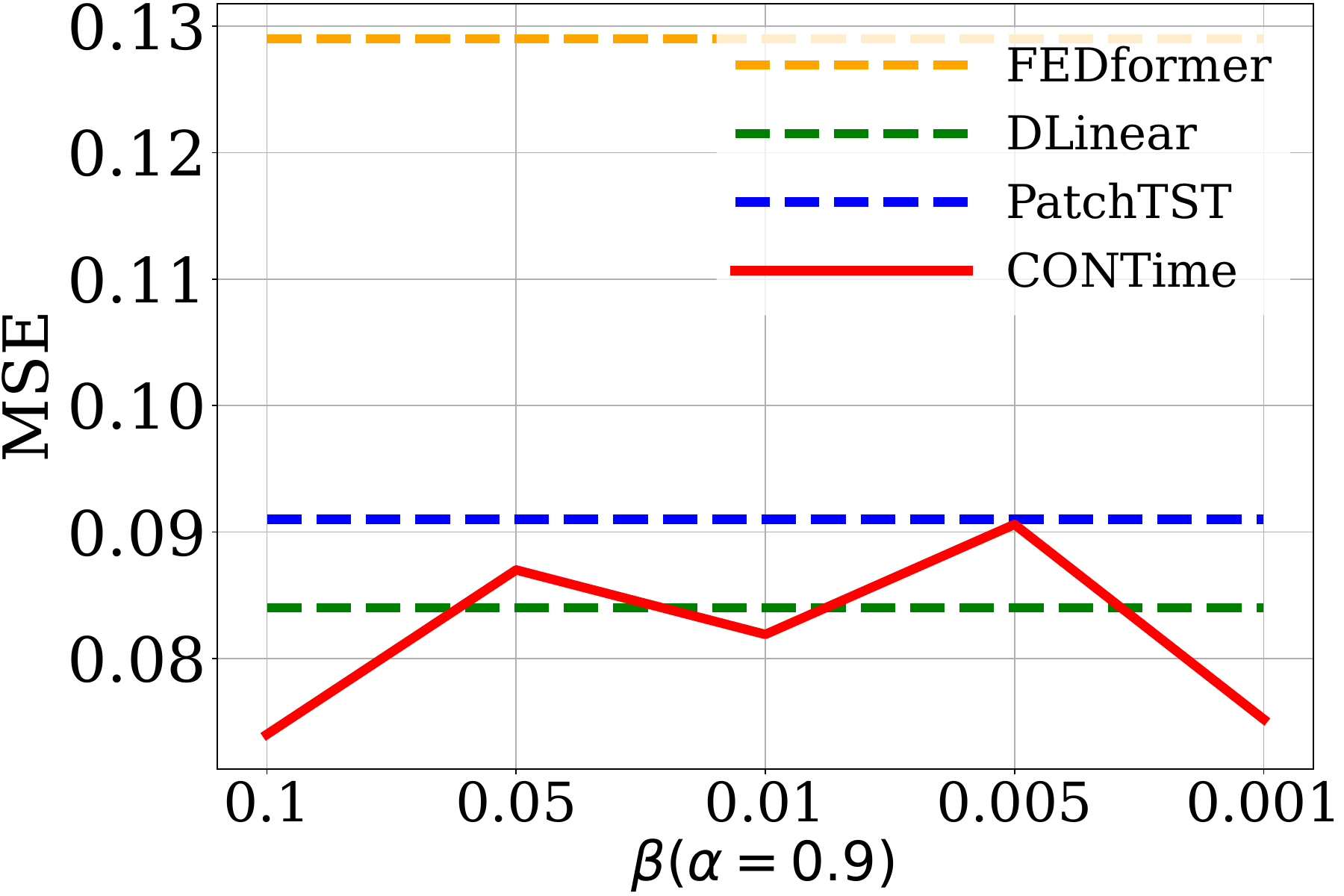}} 
\caption{Sensitivity to $\alpha, \beta$ in AAPL}
\label{fig:abl_1}
\end{center}
\vskip -0.1in
\end{figure*}

\subsection{Sensitivity analysis \& ablation study}

\subsubsection{Ablation study on loss function}

Table~\ref{tbl:abl_2} summarizes the results of the ablation study for the loss functions applied to CONTIME. Three types of loss functions are utilized. $L_{Task}$ is a plain MSE loss function to compare $Y$ and $\hat{Y}$. $L_{TDI}$ uses the TDI-based regularization proposed in~\citep{le2019shape} to address the prediction delay problem. This regularization minimizes TDI using Equation~\eqref{eq:tdi} of Appendix~\ref{appendix:TDI}. The $\Delta t$ loss function aims to reduce the computed time-derivative explicitly. Through this ablation study, we evaluate the effectiveness of the time-derivative regularization process in alleviating prediction delays. 

CONTIME (Only $L_{task}$), which trains with the task loss, exhibits reasonable performance in terms of MSE. However, it becomes apparent that it does not effectively learn in other aspects such as TDI and DTW. On the other hand, $L_{TDI}$ and $L_{\Delta t}$, employing different types of regularization respectively, demonstrate exceptional performance in terms of TDI. Though CONTIME ($L_{task}$ + $L_{TDI}$) performs well on TDI, it exhibits unstable performance in terms of MSE or DTW. However, CONTIME ($L_{task}$ + $L_{\Delta t}$) excels the others in all three evaluation metrics, proving efficacy of our $\Delta t$ loss.
\paragraph{Relationship between $L_{TDI}$ and $L_{\Delta t}$}
The experimental results in Table~\ref{tbl:abl_2} of the paper show that CONTIME ($L_{\Delta t}$) has superior TDI values compared to CONTIME ($L_{TDI}$). The TDI loss, calculated from the optimal DTW path $A$ between $\hat{y}$ and $y$, exhibits good performance in TDI, which is slghtly inferior to our methodology. Unlike the TDI loss, solely focusing on aligning the timing of the DTW path, our methodology improves performance on both DTW (shape) and TDI (timing) through the explicit gradient modeling at each time $t$.

\subsubsection{Sensitivity to $\alpha,\beta$}
In Figure~\ref{fig:abl_1}, we discern the impact of our $\Delta t$ loss on TDI, DTW, and MSE with the sensitivity curve w.r.t $\beta$ (varying from $0.1$ to $0.001$), compared to the top 3 baselines in each metric. Across all settings, CONTIME consistently outperforms the baselines in terms of DTW and TDI, demonstrating the efficacy of our model. Regarding MSE, CONTIME exhibits reasonable performance across all settings. 
These results indicate stable performance of our model trained with $\Delta t$ loss, thereby leading to more effective elimination of prediction delays and it also shows stable performance in DTW and MSE.
\subsubsection{Additional experiments on other 3 datasets}
To evaluate the performance of our model in different domains, we evaluate the model on ILL (National Disease) and ETTh1 and ETTh2. Compared to PatchTST and DLinear, we show slightly better performance in terms of MSE (value), but excellent performance in TDI and DTW metrics that measure timing and shape.
\begin{table}[t]
\centering 
\scriptsize
\caption{Additional experimental results on 3 datasets}\label{tbl:abl_3}
\begin{tabular}{ccccccccccccccccccc} \toprule
\multirow{2}{*}{Datasets} & \multirow{2}{*}{$P$}  & \multicolumn{3}{c}{CONTIME}  & \multicolumn{3}{c}{PatchTST}  & \multicolumn{3}{c}{DLinear} \\  \cmidrule(lr){3-5} \cmidrule(lr){6-8} \cmidrule(lr){9-11}
                         &  & TDI      & DTW      & MSE  & TDI      & DTW      & MSE   & TDI      & DTW      & MSE     \\\midrule 
\multirow{4}{*}{\begin{tabular}[c]{@{}c@{}}ILL\end{tabular}}
& 24   & 1.552 & 4.122 & 1.357 & 1.744 & 4.311 & 1.449 & 2.013 & 4.278 & 1.980 \\
& 36   & 1.739 & 5.108 & 1.501 & 1.927 & 5.258 & 1.541 & 3.124 & 6.018 & 1.873\\
& 48   & 1.042 & 4.916 & 0.778 & 1.224 & 5.986 & 1.673 & 4.171 & 9.701 & 2.296 \\
& 60   & 2.404 & 6.626 & 1.688 & 2.421 & 8.376 & 1.549 & 5.281 & 8.240 & 2.334 \\ \midrule
\multirow{4}{*}{\begin{tabular}[c]{@{}c@{}}ETTh1\end{tabular}}
& 24   & 1.199 & 2.235 & 0.445 & 1.921 & 2.204 & 0.329 & 1.709 & 2.907 & 0.398 \\
& 36   & 1.443 & 2.389 & 0.441 & 2.205 & 2.426 & 0.364 & 2.231 & 3.294 & 0.388 \\
& 48   & 1.872 & 2.832 & 0.437 & 2.475 & 2.758 & 0.338 & 2.621 & 3.907 & 0.378 \\
& 60   & 2.201 & 4.008 & 0.453 & 2.674 & 4.071 & 0.354 & 2.988 & 4.227 & 0.386  \\ \midrule
\multirow{4}{*}{\begin{tabular}[c]{@{}c@{}}ETTh2\end{tabular}} 
& 24   & 1.952 & 1.571 & 0.177 & 2.162 & 2.319 & 0.187 & 2.266 & 2.359 & 0.181      \\
& 36   & 2.231 & 1.957 & 0.199 & 2.479 & 2.675 & 0.202 & 2.509 & 2.883 & 0.217      \\
& 48   & 2.417 & 2.087 & 0.219 & 2.537 & 2.014 & 0.246 & 3.120 & 2.912 & 0.228       \\
& 60   & 3.370 & 2.481 & 0.233 & 3.485 & 2.528 & 0.271 & 3.839 & 3.387 & 0.263        \\\bottomrule
\end{tabular}
\end{table}

\begin{table}[t]
\centering 
\footnotesize
\caption{Comparison between CONTIME and CONTIME (shift term)}\label{tbl:abl_4}
\begin{tabular}{cccccccccccccccc} \toprule
\multirow{2}{*}{Models} & \multirow{2}{*}{$P$}  & \multicolumn{3}{c}{AAPL}  & \multicolumn{3}{c}{Weather}  \\  \cmidrule(lr){3-5} \cmidrule(lr){6-8}
                         &  & TDI      & DTW      & MSE     & TDI      & DTW      & MSE     \\\midrule 
\multirow{4}{*}{\begin{tabular}[c]{@{}c@{}}CONTIME\end{tabular}}
& 24   & 2.378 & 1.114 & 0.074 & 2.323 & 1.115 & 0.129\\
& 36   & 4.807 & 1.541 & 0.089 & 4.094 & 1.563 & 0.154\\
& 48   & 7.338 & 1.941 & 0.112 & 6.273 & 2.043 & 0.183\\
& 60   & 7.932 & 2.625 & 0.147 & 9.243 & 2.253 & 0.194\\ \midrule
\multirow{4}{*}{\begin{tabular}[c]{@{}c@{}}CONTIME\\ ($\texttt{shift term}$)\end{tabular}}
& 24   & 2.917 & 1.111 & 0.074 & 2.254 & 1.023 & 0.117 \\
& 36   & 4.802 & 1.602 & 0.094 & 4.120 & 1.390 & 0.136\\
& 48   & 7.300 & 1.912 & 0.114 & 6.226 & 1.805 & 0.159\\
& 60   & 7.964 & 2.625 & 0.148 &9.366 & 2.121 & 0.174 \\ \bottomrule
\end{tabular}
\end{table}

\subsubsection{To deal with distribution shift problem}
The most difficult part of predicting time series benchmarked datasets is the distribution shift problem~\cite{kim2021reversible}. In this paper, we use the shift method as in NLinear~\cite{zeng2023transformers} to solve this situation.
\begin{align}
\begin{split}
    \texttt{shift term}= \hat{Y}(0) - \mathbf{X}(T),\\
    \hat{Y} = \hat{Y} + \texttt{shift term},
    \end{split}
\end{align}, where $\mathbf{X}(T)$ refers to the last observations of the input sequences. We calculate the difference between the last observation $\mathbf{X}(T)$ and the first prediction value $\hat{Y}(0)$. By simply adding $\texttt{shift term}$ to the forecasting result $\hat{Y}$, We can reduce the distribution shift problem,
\section{Conclusions}

This paper suggests yet another view on the time series forecasting research in other perspectives. To mitigate the prediction latency in time series forecasting, we suggest CONTIME, a unique architecture that enables the explicit supervision of the time-derivative of observations in the continuous time domain by continuously generalizing the bi-directional GRU. With this distinctive architecture, we effectively addressed the prediction delay problem, which has long been an obstacle of time series forecasting. By applying the continuous bi-directional GRU and $\Delta t$ loss to naturally supervise the time-derivative, CONTIME alleviates the prediction delay problem. We quantify these phenomena by measuring not just MSE but also TDI and DTW as evaluation metrics. As a result, we exhibit superior overall performance when compared to 6 state-of-the-art baselines for 6 datasets from various fields.

\section*{Acknowledgements}
This work was partly supported by the Korea Advanced Institute of Science and Technology (KAIST) grant funded by the Korea government (MSIT) (No. G04240001, Physics-inspired Deep Learning, 10\%), and Institute for Information \& Communications Technology Planning \& Evaluation (IITP) grants funded by the Korea government (MSIT) (No. RS-2020-II201361, Artificial Intelligence Graduate School Program (Yonsei University),5\%), 
and (No.RS-2022-II220113, Developing a Sustainable Collaborative Multi-modal Lifelong Learning Framework,80\%) and Institute for Information \& Communications Technology Promotion (IITP) grant funded by the Korea government (MSIT) (No.RS-2019-II190075 Artificial Intelligence Graduate School Program(KAIST), 5\%)

\clearpage

\bibliographystyle{ACM-Reference-Format}
\balance
\bibliography{sample-base}

\appendix

\section{Derivatives of $\mathbf{z}(t), \mathbf{g}(t), \mathbf{r}(t)$}\label{appendix:derivatives}

\paragraph{time-derivative of $\mathbf{z}(t)$:}
The \emph{continuous} update gate is written as $\mathbf{z}(t) = \sigma\big(\mathbf{W}_z \mathbf{x}(t) + \mathbf{U}_z \mathbf{h}(t-\tau) + \mathbf{b}_z \big) = \sigma(\mathbf{A}(t, t-\tau))$, and its derivative, denoted ${\frac{ d\mathbf{z}(t)}{dt}}$, is as follows: 

\begin{align}\label{eq:derivations_z}
\begin{split}
    \frac{d\mathbf{z}(t)}{dt} = \sigma\big(\mathbf{A}(t,t-\tau))(1-\sigma(\mathbf{A}(t,t-\tau))\big) \frac{d\mathbf{A}(t,t-\tau)}{dt},
\end{split}\end{align}     
 where $\mathbf{A}(t,t-\tau) = \mathbf{W}_z \mathbf{x}(t) + \mathbf{U}_z \mathbf{h}(t-\tau) + \mathbf{b}_z$, and $\frac{d\mathbf{A}(t,t-\tau)}{dt} = \mathbf{W}_z \frac{d\mathbf{x}(t)}{dt} + \mathbf{U}_z \frac{d\mathbf{h}(t-\tau)}{dt}$.

\paragraph{time-derivative of $\mathbf{g}(t)$:}
The \emph{continuous} update vector has the form of $\mathbf{g}(t) = \phi\big(\mathbf{W}_g \mathbf{x}(t) + \mathbf{U}_g\big(\mathbf{r}(t) \odot \mathbf{h}(t-\tau)\big) + \mathbf{b}_g \big) =\phi(\mathbf{B}(t,t-\tau)$, and its derivative, ${\frac{ d\mathbf{g}(t)}{dt}}$, can be calculate as follows: 

\begin{align}\label{eq:derivations_g}
\begin{split} 
&\frac{d\mathbf{g}(t)}{dt} = \big(1-\phi^2(\mathbf{B}(t,t-\tau)\big)\frac{d\mathbf{B}(t,t-\tau)}{dt},
\end{split}\end{align} 
where $\mathbf{B}(t,t-\tau) = \mathbf{W}_g \mathbf{x}(t) + \mathbf{U}_g\big(\mathbf{r}(t) \odot \mathbf{h}(t-\tau)\big) + \mathbf{b}_g$, and $\frac{d\mathbf{B}(t,t-\tau)}{dt} = \mathbf{W}_g \frac{d\mathbf{x}(t)}{dt} + \mathbf{U}_g \frac{d\mathbf{r}(t)}{dt}\mathbf{h}(t-\tau)+\mathbf{U}_g \mathbf{r}(t)\frac{d\mathbf{h}(t-\tau)}{dt}$.

\paragraph{time-derivative of $\mathbf{r}(t)$:}
The \emph{continuous} reset gate is defined as $\mathbf{r}(t) = \sigma\big(\mathbf{W}_r \mathbf{x}(t) + \mathbf{U}_r \mathbf{h}(t-\tau) + \mathbf{b}_r \big)$, and its derivative ${\frac{ d\mathbf{r}(t)}{dt}}$ is derived as follows:

\begin{align}\label{eq:derivations_r}
\begin{split}
\frac{d\mathbf{r}(t)}{dt} = \sigma\big(\mathbf{C}(t))(1-\sigma(\mathbf{C}(t,t-\tau))\big) \frac{d\mathbf{C}(t,t-\tau)}{dt},\\
\end{split}\end{align}    
 where $\mathbf{C}(t,t-\tau) = \mathbf{W}_r \mathbf{x}(t) + \mathbf{U}_r \mathbf{h}(t-\tau) + \mathbf{b}_r$, and $\frac{d\mathbf{C}(t,t-\tau)}{dt} = \mathbf{W}_r \frac{d\mathbf{x}(t)}{dt} + \mathbf{U}_r \frac{d\mathbf{h}(t-\tau)}{dt}$.

\section{Proof of Equation.\ref{eq:derivation_full}}\label{appendix:equation_cont-GRU}
First, let $\mathbf{z}(t)$, $\mathbf{g}(t)$, and $\mathbf{r}(t)$ be the update gate, the update vector, and the reset gate of GRU: 
\begin{align}\begin{split}\label{eq:GRU_app}
\mathbf{z}(t) &= \sigma\big(\mathbf{W}_z \mathbf{x}(t) + \mathbf{U}_z \mathbf{h}(t-\tau) + \mathbf{b}_z \big),\\
\mathbf{g}(t) &= \phi\big(\mathbf{W}_g \mathbf{x}(t) + \mathbf{U}_g \big( \mathbf{r}(t) \odot \mathbf{h}(t-\tau) \big) + \mathbf{b}_g \big),\\
\mathbf{r}(t) &= \sigma\big(\mathbf{W}_r \mathbf{x}(t) + \mathbf{U}_r \mathbf{h}(t-\tau) + \mathbf{b}_r \big),
\end{split}\end{align}
To simplify the equations, we will define them as follows:

\begin{align}\begin{split}
    \mathbf{z}(t) &= \sigma\big(\mathbf{A}(t,t-\tau)\big),\\
    \mathbf{g}(t) &= \phi\big(\mathbf{B}(t,t-\tau)\big),\\
    \mathbf{r}(t) &= \sigma\big(\mathbf{C}(t,t-\tau)\big), 
\end{split}\end{align}where $\mathbf{A}(t,t-\tau) = \mathbf{W}_z \mathbf{x}(t) + \mathbf{U}_z \mathbf{h}(t-\tau) + \mathbf{b}_z$, $\mathbf{B}(t,t-\tau) = \mathbf{W}_h \mathbf{x}(t) + \mathbf{U}_h\big(\mathbf{r}(t) \odot \mathbf{h}(t-\tau)\big) + \mathbf{b}_h$, and $\mathbf{C}(t,t-\tau) = \mathbf{W}_r \mathbf{x}(t) + \mathbf{U}_r \mathbf{h}(t-\tau) + \mathbf{b}_r$. 
The derivatives of $\mathbf{z}(t)$, $\mathbf{g}(t)$, and $\mathbf{r}(t)$ are defined as follows:

\begin{align}
    \begin{split}
        \frac{d\mathbf{z}(t)}{dt} &= \sigma(\mathbf{A}(t,t-\tau))(1-\sigma(\mathbf{A}(t,t-\tau)))\frac{d\mathbf{A}(t,t-\tau)}{dt}\\
        \frac{d\mathbf{g}(t)}{dt} &= (1-\phi^2(\mathbf{B}(t,t-\tau)))\frac{d\mathbf{B}(t,t-\tau)}{dt}\\
        \frac{d\mathbf{r}(t)}{dt} &= \sigma(\mathbf{C}(t,t-\tau))(1-\sigma(\mathbf{C}(t,t-\tau)))\frac{d\mathbf{C}(t,t-\tau)}{dt}
    \end{split}
\end{align}
Lastly, the hidden state $\mathbf{h}(t)$ of GRU is written as follows: \begin{align}\label{eq:GRU_hidden}\begin{split}
\mathbf{h}(t) &= \mathbf{z}(t) \odot \mathbf{h}(t-\tau) + (1-\mathbf{z}(t)) \odot \mathbf{g}(t).
\end{split}\end{align}

The derivative of the hidden state $\mathbf{h}(t)$ is defined by the chain rule as follows: 
\begin{align}\begin{split}
    \frac{d\mathbf{h}(t)}{dt} &= \frac{d\mathbf{z}(t)}{dt} \odot \mathbf{h}(t-\tau) + \mathbf{z}(t)\odot\frac{d\mathbf{h}(t-\tau)}{dt} \\ &-\frac{d\mathbf{z}(t)}{dt}\odot \mathbf{g}(t) +(1-\mathbf{z}(t))\odot\frac{d\mathbf{g}(t)}{dt},\\
    &= \frac{d\mathbf{z}(t)}{dt} \odot \big(\mathbf{h}(t-\tau) - \mathbf{g}(t)\big) \\ &+ \mathbf{z}(t)\odot\big(\frac{d\mathbf{h}(t-\tau)}{dt} -\frac{d\mathbf{g}(t)}{dt}\big) + \frac{d\mathbf{g}(t)}{dt},\\
&= \frac{d\mathbf{z}(t)}{dt}\odot \mathbf{\zeta}(t,t-\tau) + \mathbf{z}(t)\odot \frac{d\mathbf{\zeta}(t,t-\tau)}{dt} + \frac{d\mathbf{g}(t)}{dt},\\
    \end{split}
\end{align} where $\mathbf{\zeta}(t,t-\tau) = \mathbf{h}(t-\tau)-\mathbf{g}(t)$. So, we can rewrite $\frac{d\mathbf{h}(t)}{dt}$ as follows:

\begin{align}\begin{split}
    \frac{d\mathbf{h}(t)}{dt} &= \frac{d(\mathbf{z}(t)\odot \mathbf{\zeta}(t,t-\tau))}{dt}+\frac{d\mathbf{g}(t)}{dt}
\end{split}
\end{align}

\section{Detailed descriptions of interpolation methods}\label{appendix:Interpolation}
In Section.~\ref{sec:bi_d_contime}, we calculated $\frac{dX(t)}{dt}$ by using Cubic Hermite spline method. In this section, we describe why we choose the Cubic Hermite spline method not the Natural cubic spline which creates the continuous path $X(t)$. There are two interpolation methods that create continuous path $X(t)$, Natural cubic splines and Cubic Hermite splines. 
\paragraph{Natural cubic splines:} Natural cubic splines used in Neural CDE~\cite{kidger2020neural} require the entire time series to be used as a control signal. That is, a change in future time step may interfere past time steps, thereby making interpolated result unreliable. In other words, it is an interpolation method that cannot be used in online prediction.

\paragraph{Cubic Hermite splines:} This approach mitigates the discontinuity of linear control while maintaining the same online properties by joining adjacent viewpoints with cubic splines that use additional degrees of freedom to smooth out gradient discontinuities. This results in faster integration times than linear control. The main difference from natural cubic splines is that Cubic Hermite splines solve a single equation for each $[i,i+1)$ piece independently. As a result, it changes more quickly than the natural cubic spline and therefore has a slower integration time than the natural cubic spline~\cite{morrill2021neural}.

Due to the above two differences, we believe that the Cubic Hermite spline is more suitable for real-world time series forecasting, so we use this method to create a continuous path $X(t)$.

\section{Datasets}\label{appendix:datasets}

The datasets used in our experiments are publicly available and can be downloaded at the following locations:
\begin{enumerate}
    \item AAPL: \url{https://finance.yahoo.com/quote/AAPL/history?p=AAPL},
    \item AMZN: \url{https://finance.yahoo.com/quote/AMZN/history?p=AMZN},
    \item MSFT: \url{https://finance.yahoo.com/quote/MSFT/history?p=MSFT},
    \item GOOG: \url{https://finance.yahoo.com/quote/GOOG/history?p=GOOG},
    \item Exchange: \url{https://drive.google.com/drive/folders/1ZOYpTUa82_jCcxIdTmyr0LXQfvaM9vIy},
    \item Weather: \url{https://drive.google.com/drive/folders/1ZOYpTUa82_jCcxIdTmyr0LXQfvaM9vIy},
\end{enumerate}
We split the entire dataset into training/validating/testing parts. The first 70\% of the data is used as training, 10\% is used for validating, and the last 20\% is used for testing. 

\section{Hyperparameter}\label{appendix:hyperparameter}All of the models follow the same experimental setup with prediction horizon $P \in \{24,36,48,60\}$ for all 6 datasets.
\subsection{Hyperparameter for CONTIME}


In Table~\ref{tbl:appendix_best}, we showed our best hyperparameter for all 6 datasets.

\begin{table}[h!]
\scriptsize
\caption{Best hyperparameter for CONTIME}
    \centering
    \begin{tabular}{cccccccc}\toprule
        Hyperparameter &$P$& AAPL & AMZN & GOOG & MSFT & Exchange & Weather \\ \midrule
    \multirow{4}{*}{\begin{tabular}[c]{@{}c@{}}$\lambda$\end{tabular}}
& 24   & 0.005 & 0.005 & 0.005 & 0.005 & 0.01 & 0.001\\
& 36   & 0.005 & 0.001 & 0.005 & 0.001 & 0.005 & 0.001\\
& 48   & 0.005 & 0.005 & 0.005 & 0.005 & 0.005 & 0.001 \\
& 60   & 0.005 & 0.005 & 0.005 & 0.005 & 0.005 & 0.001 \\ \midrule       
\multirow{4}{*}{\begin{tabular}[c]{@{}c@{}}$\alpha$\end{tabular}}
& 24   & 0.9 & 0.8 & 0.8 & 0.9 & 0.9 & 0.9\\
& 36   & 0.8 & 0.8 & 0.8 & 0.8 & 0.9 & 0.9\\
& 48   & 0.9 & 0.9 & 0.8 & 0.9 & 0.9 & 0.9 \\
& 60   & 0.9 & 0.9 & 0.8 & 1.0 & 0.9 & 0.9\\ \midrule    
\multirow{4}{*}{\begin{tabular}[c]{@{}c@{}}$\beta$\end{tabular}}
& 24   & 0.1 & 0.1 & 0.05 & 0.001 & 0.1 & 0.001 \\
& 36   & 0.01 & 0.01 & 0.01 & 0.001 & 0.1 & 0.001\\
& 48   & 0.001 & 0.1 & 0.05 & 0.001 & 0.1 & 0.001 \\
& 60   & 0.1 & 0.005 & 0.1 & 0.001 & 0.01 & 0.001 \\ \midrule 
\multirow{4}{*}{\begin{tabular}[c]{@{}c@{}}$T$ \end{tabular}}
& 24   & 144 & 104 & 144 & 144 & 60 & 60\\
& 36   & 144 & 144 & 144 & 104 & 60 & 60\\
& 48   & 144 & 144 & 144 & 104 & 60 & 60\\
& 60   & 144 & 104 & 144 & 144 & 60 & 60\\
\midrule 
    \end{tabular}
    \label{tbl:appendix_best}
\end{table}

\section{How to calculate TDI} \label{appendix:TDI}

We adopted TDI loss calculation for time-series sequence from \citep{le2019shape}. The calculation below is applied to each feature of data $X$ defined in Section.~\ref{sec:bi_d_contime}.

We define $\mathbf{A}$ as a binary warping path of prediction length $P$, i.e. $\mathbf{A} \subset \{0, 1\}^{P \times P}$, with $A_{h,j} = 1$ if $\hat{Y}_h$ is associated with $Y_j$ and otherwise 0, where $h, j$ are a time point of each sequence.
$\Delta(\hat{Y}, Y) := \left[(\hat{Y}_h-Y_j)^2\right]_{h,j}$, which means measuring dissimilarity of two sequences by euclidean distance. 
We calculate TDI from the optimal path matrix $\mathbf{A}^\star$ of DTW as follows:
\begin{align}
        \textit{DTW}(\hat{Y}_i, Y_i) &= \min_{\mathbf{A}\in A_{P,P}}\langle\mathbf{A}, \Delta(\hat{Y}_i, Y_i)\rangle \label{eq:dtw}
\end{align}
\begin{align}
      \mathbf{A}^\star &:= \arg\min_{\mathbf{A}\in A_{P,P}}\langle\mathbf{A}, \Delta(\hat{Y}_i, Y_i)\rangle \label{eq:a_star}
\end{align}
\begin{align}
        \textit{TDI}(\hat{Y}_i, Y_i) &:= \langle\mathbf{A}^\star, \Omega\rangle \label{eq:tdi} = \left\langle \arg\min_{\mathbf{A}\in A_{p,p}}\langle\mathbf{A}, \Delta(\hat{Y}_i, Y_i)\rangle, \Omega\right\rangle
\end{align} where $\Omega$ is a square matrix of size $P \times P$ penalizing each element $Y_h$ being associated to an $\hat{Y}_j$, for $h \ne j$ : e.g. $\Omega(h, j) = \frac{(h-j)^2}{P^2}$. \\
To make TDI differentiable, we approximate $\mathbf{A}^\star$ with $\mathbf{A}_\gamma^\star$ using the fact that $\mathbf{A}^\star = \nabla_\Delta \textit{DTW}(\hat{Y}_i, Y_i)$:
\begin{align}\label{eq:a_star_}
        \mathbf{A}^\star_\gamma &:= \nabla_\Delta \textit{DTW}_{\gamma} (\hat{Y}_i, Y_i) = \frac{1}{Z} \sum_{\mathbf{A}\in A_{P,P}}\mathbf{A}\exp{-{\frac{\langle\mathbf{A}, \Delta(\hat{Y}_i, Y_i)\rangle}{\gamma}}}
\end{align} 
where $Z=\exp{-{\frac{\langle\mathbf{A}, \Delta(\hat{Y}_i, Y_i)\rangle}{\gamma}}}$. 
The resulting TDI loss is:

\begin{align}
L_{TDI} := \textit{TDI}(\hat{Y}_i, Y_i) := \langle\mathbf{A}^\star_\gamma, \Omega\rangle \label{eq:tdi}
\end{align}

\section{Forecasting Visualization} \label{appendix:forecasting_visualization}
In this section, we additionally visualize forecasting results on all 6 datasets. 
\begin{figure}
    \centering
    \subfigure[GOOG]{\includegraphics[width=1.0\columnwidth]{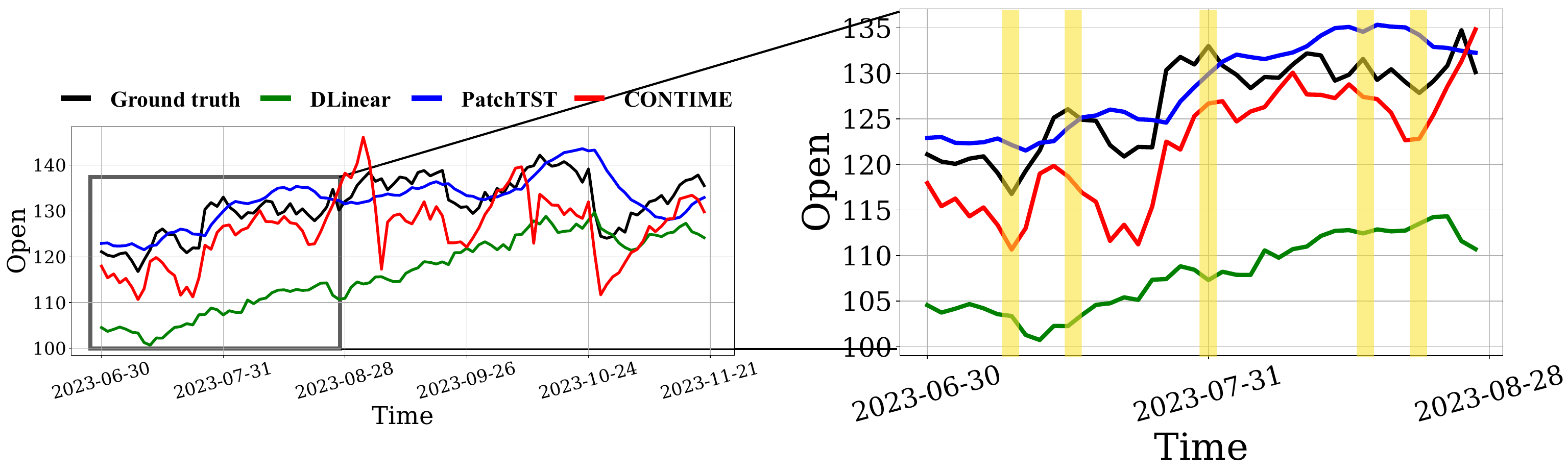}}\\
    \subfigure[MSFT]{\includegraphics[width=1.0\columnwidth]{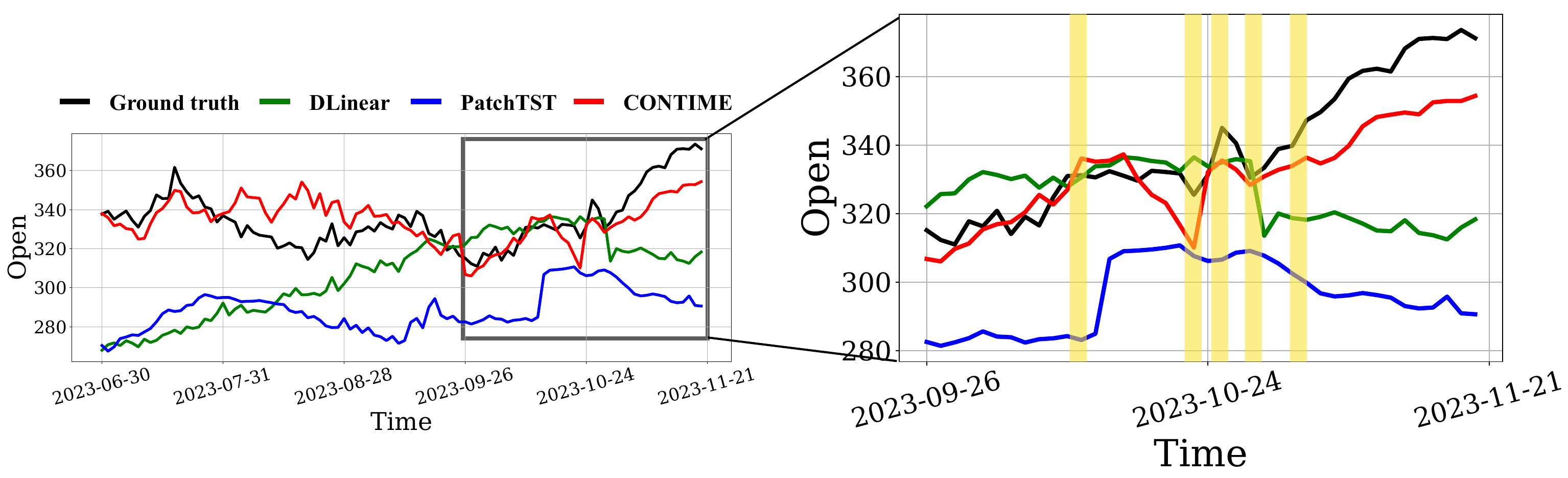}}\\
    \caption{Forecasting visualization on 2 datasets}
    \label{fig:appendix_forecasting_vis1}
\end{figure}

    
\section{Computational time and model usage}\label{appendix:time_model}
In this section, we report computational time of our model and model usage for all 6 datasets. 
\begin{table}[h!]
\caption{Computational time}
\scriptsize
    \centering
    \begin{tabular}{ccccccc}\toprule
        Models & AAPL & AMZN & GOOG & MSFT & Exchange & Weather \\ \midrule
    
DLinear & 6.436 & 5.653 & 6.289 & 6.371 & 41.25 & 240.9\\
NODE   & 14.08 & 13.85 & 13.67 & 13.81 & 43.40 & 9.935\\
NCDE   & 80.68 & 82.72 & 77.69 & 36.52 & 34.94 & 86.29 \\
Autoformer   & 8.414 & 7.949 & 8.704 & 8.191 & 31.99 & 350.2 \\        
FEDformer   & 14.58 & 10.03 & 9.598 & 11.20 & 39.18 & 312.83\\
PatchTST   & 2.495 & 2.240 & 2.751 & 2.236 & 22.62 & 289.6 \\ \midrule
CONTIME  & 23.25 & 21.79 & 23.65 & 32.37 & 33.68 & 29.93 \\ \toprule
    \end{tabular}
    \label{tbl:time_complexity}
\end{table} 

\begin{table}[h!]
\caption{Model usage}
\scriptsize
    \centering
    \begin{tabular}{ccccccc}\toprule
        Models & AAPL & AMZN & GOOG & MSFT & Exchange & Weather \\ \midrule
    
DLinear & 179.1 & 179.1 & 179.1 & 179.1 & 173.2 & 180.7\\
NODE   & 26.05 & 26.05 & 26.05 & 26.05 & 27.37 & 35.39\\
NCDE   & 7.027 & 7.027 & 7.027 & 7.027 & 10.78 & 50.69 \\
Autoformer   & 112.4 & 112.4 & 112.4 & 112.4 & 1,243 & 1,244 \\        
FEDformer   & 496.7 & 496.7 & 496.7 & 496.7 & 635.1 & 804.1\\
PatchTST   & 102.3 & 102.3 & 102.3 & 102.3 & 116.8 & 625.2 \\ \midrule
CONTIME  & 198.7 & 198.7 & 198.7 & 198.7 & 117.7 & 131.8 \\ \toprule
    \end{tabular}
    \label{tbl:time_complexity}
\end{table} 


\section{Hyperparameter}\label{appendix:hyperparameter}All of the models follow the same experimental setup with prediction horizon $P \in \{24,36,48,60\}$ for all 6 datasets.
\subsection{Hyperparameter for baselines} 
For the best outcome of baselines and our method, we conduct a hyperparameter search for them based on the recommended hyperparameter set from each paper. Considered hyperparameter sets are as follows:

\paragraph{Stocks:}
\begin{enumerate}
    \item DLinear : We train for 100 epochs with a learning rate $\lambda$ in $\{0.01,0.05, 0.001,0.005,0.0001\}$. Input sequence length $T$ in $\{96,104,144\}$
    \item Differential-equation based models: For Neural ODE and Neural CDE, we train for 100 epochs with a learning rate $\lambda$ in $\{0.01,0.05, 0.001,0.005,0.0001\}$. Hidden size in $\{39,49,59\}$.
    \item Transformer-based models: For Autoformer and FEDformer, we train for 50 epochs with a learning rate $\lambda$ in $\{0.01,0.05,\\0.001,0.005,0.0001\}$. Input sequence length $T$ in $\{96,104,144\}$
    \item PatchTST: We train for 50 epochs with a learning rate $\lambda$ in $\{0.01,0.05,0.001,0.005,0.0001\}$. Input sequence length $T$ in $\{96,104,144\}$. Other settings follow the same experimental settings in the baseline.
\end{enumerate}

\paragraph{Exchange:}
\begin{enumerate}
    \item DLinear : We train for 100 epochs with a learning rate $\lambda$ in $\{0.01,0.05, 0.001,0.005,0.0001\}$. Input sequence length $T$ in $\{60,72,96\}$.
    \item Differential-equation based models: For Neural ODE and Neural CDE, we train for 100 epochs with a learning rate $\lambda$ in $\{0.01,0.05, 0.001,0.005,0.0001\}$. Hidden size in $\{39,49,59\}$. Input sequence length $T$ in $\{60,72,96\}$.
    \item Transformer-based models: For Autoformer and FEDformer, we train for 50 epochs with a learning rate $\lambda$ in $\{0.01,0.05,\\0.001,0.005,0.0001\}$. Input sequence length $T$ in $\{60,72,96\}$. Other settings follow the same experimental settings in the baseline.
    \item PatchTST: We train for 50 epochs with a learning rate $\lambda$ in $\{0.01,0.05,0.001,0.005,0.0001\}$. Input sequence length $T$ in $\{60,72,96\}$. Other settings follow the same experimental settings in the baseline.
\end{enumerate}

\paragraph{Weather:}
\begin{enumerate}
    \item DLinear : We train for 100 epochs with a learning rate $\lambda$ in $\{0.01,0.05, 0.001,0.005,0.0001\}$. Input sequence length $T$ in $\{96,104,144\}$
    \item Differential-equation based models: For Neural ODE and Neural CDE, we train for 100 epochs with a learning rate $\lambda$ in $\{0.01,0.05, 0.001,0.005,0.0001\}$. Hidden size in $\{39,49,59\}$. Input sequence length $T$ in $\{60,72,96\}$.
    \item Transformer-based models: For Autoformer and FEDformer, we train for 50 epochs with a learning rate $\lambda$ in $\{0.01,0.05,\\0.001,0.005,0.0001\}$. Input sequence length $T$ in $\{60,72,96\}$. Other settings follow the same experimental settings in the baseline.   
    \item PatchTST: We train for 50 epochs with a learning rate $\lambda$ in $\{0.01,0.05,0.001,0.005,0.0001\}$. Input sequence length $T$ in $\{60,72,96\}$. Other settings follow the same experimental settings in the baseline.   
\end{enumerate}

\subsection{Hyperparameter for CONTIME}
For reproducibility, we report the hyperparameters search range as follows:
\paragraph{Stocks:}
We train for 100 epochs with a batch size of $256$. A learning rate $\lambda$ in $\{0.001,0.005,0.01,0.05\}$ are used. Coefficient of $L_{CONTIME} \alpha$ in $\{0.7,0.8,0.9,1.0\}$ and $\beta$ in $\{ 0.001, 0.005, 0.01, 0.05, 0.1\}$. We used rk4 as an ODE solver. Our input sequence length $T$ in $\{96,104,144\}$.

\paragraph{Exchange:}
We train for 100 epochs with a batch size of $256$. A learning rate $\lambda$ in $\{0.001,0.005,0.01,0.05\}$ are used. Coefficient of $L_{CONTIME} \alpha$ in $\{0.7,0.8,0.9,1.0\}$ and $\beta$ in $\{ 0.001, 0.005, 0.01, 0.05, 0.1\}$. We used rk4 as an ODE solver. Our input sequence length $T$ in $\{60,72,96\}$. 
\paragraph{Weather:}
We train for 150 epochs with a batch size of $256$. A learning rate $\lambda$ in $\{0.001,0.005,0.01,0.05\}$ are used. Coefficient of $L_{CONTIME}$ $\alpha$ in $\{0.7,0.8,0.9,1.0\}$ and $\beta$ in $\{ 0.001, 0.005, 0.01, 0.05, 0.1\}$. We used rk4 as an ODE solver. Our input sequence length $T$ in $\{60,72,96\}$.

In Table~\ref{tbl:appendix_best}, we showed our best hyperparameter for all 6 datasets.

\begin{table}[h!]
\scriptsize
\caption{Best hyperparameter for CONTIME}
    \centering
    \begin{tabular}{cccccccc}\toprule
        Hyperparameter &$P$& AAPL & AMZN & GOOG & MSFT & Exchange & Weather \\ \midrule
    \multirow{4}{*}{\begin{tabular}[c]{@{}c@{}}$\lambda$\end{tabular}}
& 24   & 0.005 & 0.005 & 0.005 & 0.005 & 0.01 & 0.001\\
& 36   & 0.005 & 0.001 & 0.005 & 0.001 & 0.005 & 0.001\\
& 48   & 0.005 & 0.005 & 0.005 & 0.005 & 0.005 & 0.001 \\
& 60   & 0.005 & 0.005 & 0.005 & 0.005 & 0.005 & 0.001 \\ \midrule       
\multirow{4}{*}{\begin{tabular}[c]{@{}c@{}}$\alpha$\end{tabular}}
& 24   & 0.9 & 0.8 & 0.8 & 0.9 & 0.9 & 0.9\\
& 36   & 0.8 & 0.8 & 0.8 & 0.8 & 0.9 & 0.9\\
& 48   & 0.9 & 0.9 & 0.8 & 0.9 & 0.9 & 0.9 \\
& 60   & 0.9 & 0.9 & 0.8 & 1.0 & 0.9 & 0.9\\ \midrule    
\multirow{4}{*}{\begin{tabular}[c]{@{}c@{}}$\beta
$\end{tabular}}
& 24   & 0.1 & 0.1 & 0.05 & 0.001 & 0.1 & 0.001 \\
& 36   & 0.01 & 0.01 & 0.01 & 0.001 & 0.1 & 0.001\\
& 48   & 0.001 & 0.1 & 0.05 & 0.001 & 0.1 & 0.001 \\
& 60   & 0.1 & 0.005 & 0.1 & 0.001 & 0.01 & 0.001 \\ \midrule 
\multirow{4}{*}{\begin{tabular}[c]{@{}c@{}}$T$ \end{tabular}}
& 24   & 144 & 104 & 144 & 144 & 60 & 60\\
& 36   & 144 & 144 & 144 & 104 & 60 & 60\\
& 48   & 144 & 144 & 144 & 104 & 60 & 60\\
& 60   & 144 & 104 & 144 & 144 & 60 & 60\\
\midrule 
    \end{tabular}
    \label{tbl:appendix_best}
\end{table}

\section{Experimental results with standard deviation}\label{appendix:exp_std}
In Table~~\ref{tbl:abl_exp_1} and Table~\ref{tbl:abl_exp_2}, we repeat training and testing procedures with three different random seeds and report their mean squared error and standard deviations of all 6 datasets.
\begin{table*}[t]
\centering 
\caption{Experimental results on 3 datasets. The best results are in \textbf{bold} and the second best are \underline{underlined}.}\label{tbl:abl_exp_1}
\begin{tabular}{ccccccccccccccccccc} \toprule
\multicolumn{2}{c}{Datasets}   &  \multicolumn{3}{c}{APPL} & \multicolumn{3}{c}{AMZN} & \multicolumn{3}{c}{GOOG}  \\\cmidrule(lr){1-2} \cmidrule(lr){3-5} \cmidrule(lr){6-8} \cmidrule(lr){9-11} 
 & $$P$$   & TDI  & DTW  & MSE  & TDI & DTW & MSE  & TDI & DTW & MSE    \\\midrule
\multirow{4}{*}{\begin{sideways}DLinear\end{sideways}} 
& 24    & 3.810 \scriptsize{± 0.031} & 1.409 \scriptsize{± 0.034} & 0.105 \scriptsize{± 0.004} 
        & 3.855 \scriptsize{± 0.048} & 2.239 \scriptsize{± 0.050} & 0.265 \scriptsize{± 0.008}  
        & 3.766 \scriptsize{± 0.031} & \underline{1.297} \scriptsize{± 0.006} & \underline{0.166} \scriptsize{± 0.008} \\
& 36    & 5.106 \scriptsize{± 0.028} & 1.940 \scriptsize{± 0.049} & 0.187 \scriptsize{± 0.008} 
        & \underline{5.396} \scriptsize{± 0.051} & 2.726 \scriptsize{± 0.054} & 0.372 \scriptsize{± 0.006}
        & \underline{4.835} \scriptsize{± 0.044} & 2.229 \scriptsize{± 0.084} & 0.199 \scriptsize{± 0.008}   \\
& 48    & 7.751 \scriptsize{± 1.060} & 2.323 \scriptsize{± 0.033} & 0.213 \scriptsize{± 0.008} 
        & 8.915 \scriptsize{± 0.056} & 2.964 \scriptsize{± 0.073} & 0.408 \scriptsize{± 0.008} 
        & \underline{7.518} \scriptsize{± 0.077} & 2.568 \scriptsize{± 0.034} & 0.262 \scriptsize{± 0.003}  \\
& 60    & 10.84 \scriptsize{± 1.276} & 2.907 \scriptsize{± 0.143} & 0.258 \scriptsize{± 0.010} 
        & 9.252 \scriptsize{± 0.047} & 3.017 \scriptsize{± 0.055} & 0.347 \scriptsize{± 0.009}
        & 12.39 \scriptsize{± 0.038} & 2.848 \scriptsize{± 0.061} & 0.294 \scriptsize{± 0.009}  \\\midrule
\multirow{4}{*}{\begin{sideways}NODE\end{sideways}} 
& 24    & 3.739 \scriptsize{± 0.047} & 4.330 \scriptsize{± 0.008} & 0.168 \scriptsize{± 0.007}  
        & 3.063 \scriptsize{± 0.021} & 3.275 \scriptsize{± 0.009} & 0.397 \scriptsize{± 0.001} 
        & 3.684 \scriptsize{± 0.055} & 6.399 \scriptsize{± 0.010} & 1.298 \scriptsize{± 0.091} \\
& 36    & \underline{4.911} \scriptsize{± 0.071} & 2.916 \scriptsize{± 0.008} & 0.328 \scriptsize{± 0.008} 
        & 5.479 \scriptsize{± 0.033} & 4.893 \scriptsize{± 0.009} & 0.464 \scriptsize{± 0.001} 
        & 5.793 \scriptsize{± 0.167} & 4.223 \scriptsize{± 0.012} & 0.646 \scriptsize{± 0.001} \\
& 48    & 7.482 \scriptsize{± 0.028} & 4.203 \scriptsize{± 0.010} & 0.535 \scriptsize{± 0.001} 
        & 7.149 \scriptsize{± 0.092} & 6.436 \scriptsize{± 0.009} & 0.813 \scriptsize{± 0.087} 
        & 7.795 \scriptsize{± 0.041} & 5.112 \scriptsize{± 0.006} & 0.794 \scriptsize{± 0.001} \\
& 60    & \underline{8.702} \scriptsize{± 0.030} & 10.25 \scriptsize{± 0.007} & 1.149 \scriptsize{± 0.021} 
        & \underline{8.954} \scriptsize{± 0.073} & 6.333 \scriptsize{± 0.011} & 1.033 \scriptsize{± 0.001} 
        & \underline{9.513} \scriptsize{± 0.065} & 5.648 \scriptsize{± 0.014} & 0.874 \scriptsize{± 0.001}  \\\midrule
\multirow{4}{*}{\begin{sideways}NCDE\end{sideways}} 
& 24    & 5.039 \scriptsize{± 0.020} & 4.555 \scriptsize{± 0.033} & 0.227 \scriptsize{± 0.014} 
        & \underline{2.984} \scriptsize{± 0.037} & 5.493 \scriptsize{± 0.041} & 0.261 \scriptsize{± 0.012} 
        & 3.719 \scriptsize{± 0.046} & 4.601 \scriptsize{± 0.040} & 0.517 \scriptsize{± 0.017} \\
& 36    & 6.651 \scriptsize{± 0.032} & 3.199 \scriptsize{± 0.074} & 0.462 \scriptsize{± 0.082} 
        & 5.829 \scriptsize{± 0.036} & 4.022 \scriptsize{± 0.032} & 0.335 \scriptsize{± 0.016}
        & 4.946 \scriptsize{± 0.057} & 3.541 \scriptsize{± 0.042} & 0.582 \scriptsize{± 0.009}\\
& 48    & \underline{7.303} \scriptsize{± 0.044} & 4.028 \scriptsize{± 0.056} & 0.440 \scriptsize{± 0.019}
        & 7.113 \scriptsize{± 0.038} & 5.817 \scriptsize{± 0.063} & 0.711 \scriptsize{± 0.011}
        & 8.132 \scriptsize{± 0.053} & 6.161 \scriptsize{± 0.073} & 0.756 \scriptsize{± 0.011} \\
& 60    & 11.47 \scriptsize{± 0.093} & 3.882 \scriptsize{± 0.002} & 0.459 \scriptsize{± 0.015} 
        & 9.041 \scriptsize{± 0.043} & 7.936 \scriptsize{± 0.062} & 1.352 \scriptsize{± 0.014}
        & 10.02 \scriptsize{± 0.025} & 5.637 \scriptsize{± 0.053} & 0.771 \scriptsize{± 0.011}  \\\midrule
\multirow{4}{*}{\begin{sideways}Autoformer\end{sideways}}     
& 24    & \underline{3.085} \scriptsize{± 0.024} & 1.551 \scriptsize{± 0.023} & 0.150 \scriptsize{± 0.024}
        & 3.576 \scriptsize{± 0.037} & \textbf{1.485} \scriptsize{± 0.022} & \underline{0.174} \scriptsize{± 0.001}
        & 3.289 \scriptsize{± 0.031} & \textbf{1.239} \scriptsize{± 0.021} & 0.167 \scriptsize{± 0.001} \\
& 36    & 6.561 \scriptsize{± 0.025} & 1.882 \scriptsize{± 0.029} & 0.171 \scriptsize{± 0.001}
        & 5.541 \scriptsize{± 0.087} & 2.032 \scriptsize{± 0.018} & 0.203 \scriptsize{± 0.011}
        & 5.782 \scriptsize{± 0.061} & 2.210 \scriptsize{± 0.021} & 0.199 \scriptsize{± 0.034}\\
& 48    & 9.814 \scriptsize{± 0.031} & 2.307 \scriptsize{± 0.019} & 0.170 \scriptsize{± 0.001}
        & \underline{6.941} \scriptsize{± 0.063} & \underline{2.388} \scriptsize{± 0.013} & 0.219 \scriptsize{± 0.017}
        & 7.606 \scriptsize{± 0.040} & 2.943 \scriptsize{± 0.027} & 0.289 \scriptsize{± 0.061} \\
& 60    & 13.82 \scriptsize{± 0.052} & 2.651 \scriptsize{± 0.019} & \underline{0.188} \scriptsize{± 0.001}
        & 9.414 \scriptsize{± 0.035} & \textbf{2.723} \scriptsize{± 0.021} & \underline{0.275} \scriptsize{± 0.001}
        & 10.80 \scriptsize{± 0.021} & 3.248 \scriptsize{± 0.023} & 0.279 \scriptsize{± 0.001} \\\midrule
\multirow{4}{*}{\begin{sideways}FEDformer\end{sideways}}  
& 24    & 3.417 \scriptsize{± 0.045} & 1.396 \scriptsize{± 0.063} & 0.129 \scriptsize{± 0.009}
        & 3.108 \scriptsize{± 0.100} & 1.764 \scriptsize{± 0.056} & 0.232  \scriptsize{± 0.004} 
        & \underline{3.154} \scriptsize{± 0.034} & 1.587 \scriptsize{± 0.041} & 0.204 \scriptsize{± 0.005}\\
& 36    & 6.335 \scriptsize{± 0.031} & 1.826 \scriptsize{± 0.052} & 0.149 \scriptsize{± 0.005}
        & 5.878 \scriptsize{± 0.078} & 2.201 \scriptsize{± 0.043} & 0.249 \scriptsize{± 0.079} 
        & 5.311 \scriptsize{± 0.071} & \underline{2.203} \scriptsize{± 0.030} & 0.215 \scriptsize{± 0.004} \\
& 48    & 12.64 \scriptsize{± 0.074} & 1.932 \scriptsize{± 0.031} & 0.135 \scriptsize{± 0.004}
        & 7.664 \scriptsize{± 0.103} & 2.691 \scriptsize{± 0.071} & 0.289 \scriptsize{± 0.003} 
        & 8.489 \scriptsize{± 0.080} & \underline{2.312} \scriptsize{± 0.016} & 0.225 \scriptsize{± 0.003}\\
& 60    & 16.39 \scriptsize{± 0.081} & 2.642 \scriptsize{± 0.102} & 0.204 \scriptsize{± 0.007} 
        & 12.84 \scriptsize{± 0.108} & 2.980 \scriptsize{± 0.027} & 0.354 \scriptsize{± 0.007} 
        & 12.13 \scriptsize{± 0.081} & 2.785 \scriptsize{± 0.021} & \underline{0.244} \scriptsize{± 0.008}\\\midrule
\multirow{4}{*}{\begin{sideways}PatchTST\end{sideways}}  
& 24    & 3.166 \scriptsize{± 0.149} & \underline{1.253} \scriptsize{± 0.046} & \underline{0.084} \scriptsize{± 0.006} 
        & 3.969 \scriptsize{± 0.078} & 1.574 \scriptsize{± 0.051} & 0.177 \scriptsize{± 0.006} 
        & 3.706 \scriptsize{± 0.066} & 1.554 \scriptsize{± 0.043} & \textbf{0.165} \scriptsize{± 0.007}\\
& 36    & 5.358 \scriptsize{± 0.123} & \textbf{1.417} \scriptsize{± 0.556} & \underline{0.118} \scriptsize{± 0.006} 
        & 6.679 \scriptsize{± 0.089} & \textbf{1.733} \scriptsize{± 0.022} & \textbf{0.168} \scriptsize{± 0.008} 
        & 4.882 \scriptsize{± 0.054} & 2.766 \scriptsize{± 0.037} & \underline{0.191} \scriptsize{± 0.009} \\
& 48    & 7.984 \scriptsize{± 0.693} & \textbf{1.809} \scriptsize{± 0.459} & \underline{0.130} \scriptsize{± 0.063} 
        & 8.706 \scriptsize{± 0.062} & 2.521 \scriptsize{± 0.045} & \underline{0.220} \scriptsize{± 0.011}
        & 7.840 \scriptsize{± 0.061} & 2.342 \scriptsize{± 0.051} & \underline{0.203} \scriptsize{± 0.006}\\
& 60    & 11.00 \scriptsize{± 1.025} & \underline{2.626} \scriptsize{± 0.042} & 0.202 \scriptsize{± 0.057} 
        & 12.24 \scriptsize{± 0.042} & 3.475 \scriptsize{± 0.076} & \underline{0.275} \scriptsize{± 0.009}
        & 10.64 \scriptsize{± 0.043} & \textbf{2.673} \scriptsize{± 0.061} & \underline{0.244} \scriptsize{± 0.007}\\\bottomrule
\multirow{4}{*}{\begin{sideways}CONTIME\end{sideways}} 
& 24    & \textbf{2.378} \scriptsize{± 0.019} & \textbf{1.114} \scriptsize{± 0.013} & \textbf{0.074} \scriptsize{± 0.008} 
        & \textbf{2.866} \scriptsize{± 0.022} & \underline{1.529} \scriptsize{± 0.017} & \textbf{0.167} \scriptsize{± 0.001} 
        & \textbf{3.052} \scriptsize{± 0.011} & 1.541 \scriptsize{± 0.017} & \textbf{0.165} \scriptsize{± 0.004}   \\
& 36    & \textbf{4.807} \scriptsize{± 0.033} & \underline{1.541} \scriptsize{± 0.016} & \textbf{0.089} \scriptsize{± 0.002} 
        & \textbf{5.275} \scriptsize{± 0.008} & \underline{1.881} \scriptsize{± 0.035} & \underline{0.193} \scriptsize{± 0.090}
        & \textbf{4.712} \scriptsize{± 0.021} & \textbf{2.189} \scriptsize{± 0.010} & \textbf{0.189} \scriptsize{± 0.007}   \\
& 48    & \textbf{7.300} \scriptsize{± 0.011} & \underline{1.912} \scriptsize{± 0.008} & \textbf{0.114} \scriptsize{± 0.002} 
        & \textbf{6.844} \scriptsize{± 0.012} & \textbf{2.300} \scriptsize{± 0.016} & \textbf{0.209} \scriptsize{± 0.001} 
        & \textbf{7.364} \scriptsize{± 0.020} & \textbf{2.297} \scriptsize{± 0.016} & \textbf{0.188} \scriptsize{± 0.003}  \\
& 60    & \textbf{7.932} \scriptsize{± 0.017} & \textbf{2.625} \scriptsize{± 0.016} & \textbf{0.147} \scriptsize{± 0.009} 
        & \textbf{8.885} \scriptsize{± 0.018} & \underline{2.873} \scriptsize{± 0.013} & \textbf{0.239} \scriptsize{± 0.006}    
        & \textbf{9.271} \scriptsize{± 0.015} & \underline{2.741} \scriptsize{± 0.012} & \textbf{0.210} \scriptsize{± 0.001}  \\\bottomrule
\end{tabular}
\end{table*}

\begin{table*}
\centering 
\caption{Experimnetal results on 3 datasets.}\label{tbl:abl_exp_2}
\begin{tabular}{cccccccccccccccc} \toprule
\multicolumn{2}{c}{Datasets}   &  \multicolumn{3}{c}{MSFT} & \multicolumn{3}{c}{Exchange Rate} & \multicolumn{3}{c}{Weather} \\\cmidrule(lr){1-2} \cmidrule(lr){3-5} \cmidrule(lr){6-8} \cmidrule(lr){9-11} 
 & $$P$$   & TDI & DTW & MSE  & TDI & DTW & MSE  & TDI & DTW & MSE    \\\midrule
\multirow{4}{*}{\begin{sideways}DLinear\end{sideways}} 
& 24    & 4.327 \scriptsize{± 0.044} & \textbf{1.430} \scriptsize{± 0.091} & \underline{0.197} \scriptsize{± 0.071} 
        & 3.629 \scriptsize{± 0.052} & \textbf{0.533} \scriptsize{± 0.003} & \textbf{0.044} \scriptsize{± 0.091} 
        & 3.505 \scriptsize{± 0.071} & 1.894 \scriptsize{± 0.838} & \underline{0.119} \scriptsize{± 0.041}   \\
& 36    & 6.103 \scriptsize{± 0.083} & 2.385 \scriptsize{± 0.044} & 0.319 \scriptsize{± 0.062} 
        & 5.638 \scriptsize{± 0.088} & \underline{0.781} \scriptsize{± 0.023} & \underline{0.065} \scriptsize{± 0.071} 
        & 5.944 \scriptsize{± 0.091} & \underline{1.436} \scriptsize{± 0.064} & \underline{0.144} \scriptsize{± 0.838}  \\
& 48    & \underline{7.324} \scriptsize{± 0.087} & 3.738 \scriptsize{± 0.014} & 0.468 \scriptsize{± 0.071} 
        & 7.989 \scriptsize{± 0.852} & \underline{1.742} \scriptsize{± 0.041} & \textbf{0.084} \scriptsize{± 0.052} 
        & 8.208 \scriptsize{± 0.041} & 1.817 \scriptsize{± 0.002} & \underline{0.161} \scriptsize{± 0.064}  \\
& 60    & 12.10 \scriptsize{± 0.062} & 4.247 \scriptsize{± 0.062} & 0.492 \scriptsize{± 0.048} 
        & 11.01 \scriptsize{± 0.174} & 2.304 \scriptsize{± 0.029} & \underline{0.107} \scriptsize{± 0.064} 
        & 10.16 \scriptsize{± 0.057} & \textbf{1.771} \scriptsize{± 0.041} & \textbf{0.174} \scriptsize{± 0.014}  \\\midrule
\multirow{4}{*}{\begin{sideways}NODE\end{sideways}} 
& 24    & 4.596 \scriptsize{± 0.071} & 3.389 \scriptsize{± 0.002} & 0.359 \scriptsize{± 0.012} 
        & 2.085 \scriptsize{± 0.031} & 2.855 \scriptsize{± 0.020} & 0.525 \scriptsize{± 0.014} 
        & 2.758 \scriptsize{± 0.084} & 4.262 \scriptsize{± 0.052} & 0.336 \scriptsize{± 0.013} \\
& 36    & 6.769 \scriptsize{± 0.022} & 4.329 \scriptsize{± 0.012} & 0.496 \scriptsize{± 0.019} 
        & \underline{4.055} \scriptsize{± 0.081} & 9.289 \scriptsize{± 0.051} & 1.137 \scriptsize{± 0.011} 
        & \underline{4.314} \scriptsize{± 0.090} & 6.193 \scriptsize{± 0.031} & 0.489 \scriptsize{± 0.012} \\
& 48    & 8.868 \scriptsize{± 0.101} & 4.656 \scriptsize{± 0.030} & 0.504 \scriptsize{± 0.013} 
        & 6.104 \scriptsize{± 0.073} & 6.028 \scriptsize{± 0.031} & 1.100 \scriptsize{± 0.018} 
        & 6.827 \scriptsize{± 0.082} & 6.294 \scriptsize{± 0.022} & 1.261 \scriptsize{± 0.018} \\
& 60    & \underline{10.72} \scriptsize{± 0.073} & 7.973 \scriptsize{± 0.025} & 0.618 \scriptsize{± 0.031} 
        & 9.822 \scriptsize{± 0.091} & 6.621 \scriptsize{± 0.021} & 1.056 \scriptsize{± 0.021} 
        & 10.54 \scriptsize{± 0.080} & 7.652 \scriptsize{± 0.032} & 1.506 \scriptsize{± 0.020}  \\\midrule
\multirow{4}{*}{\begin{sideways}NCDE\end{sideways}} 
& 24    & 4.842 \scriptsize{± 0.022} & 2.809 \scriptsize{± 0.081} & 0.445 \scriptsize{± 0.064} 
        & \underline{1.874} \scriptsize{± 0.081} & 3.689 \scriptsize{± 0.092} & 0.576 \scriptsize{± 0.030} 
        & \underline{2.489} \scriptsize{± 0.031} & 5.609 \scriptsize{± 0.061} & 0.854 \scriptsize{± 0.081} \\
& 36    & 6.687 \scriptsize{± 0.082} & 2.902 \scriptsize{± 0.034} & 0.628 \scriptsize{± 0.016} 
        & 4.184 \scriptsize{± 0.045} & 8.137 \scriptsize{± 0.033} & 0.542 \scriptsize{± 0.087} 
        & 4.661 \scriptsize{± 0.027} & 4.059 \scriptsize{± 0.061} & 0.799 \scriptsize{± 0.038} \\
& 48    & 9.018 \scriptsize{± 0.071} & 4.327 \scriptsize{± 0.065} & 0.690 \scriptsize{± 0.063} 
        & \underline{6.012} \scriptsize{± 0.082} & 7.957 \scriptsize{± 0.088} & 0.874 \scriptsize{± 0.024} 
        & 6.922 \scriptsize{± 0.075} & 4.682 \scriptsize{± 0.046} & 0.783 \scriptsize{± 0.076} \\
& 60    & 12.35 \scriptsize{± 0.091} & 5.221 \scriptsize{± 0.022} & 0.766 \scriptsize{± 0.012} 
        & \underline{8.105} \scriptsize{± 0.049} & 6.516 \scriptsize{± 0.029} & 0.604 \scriptsize{± 0.094} 
        & 9.900 \scriptsize{± 0.039} & 5.882 \scriptsize{± 0.081} & 0.989 \scriptsize{± 0.053}  \\\midrule
\multirow{4}{*}{\begin{sideways}Autoformer\end{sideways}}     
& 24    & \underline{4.222} \scriptsize{± 0.041} & 1.690 \scriptsize{± 0.037} & 0.246 \scriptsize{± 0.009}           
        & 3.158 \scriptsize{± 0.158} & 1.120 \scriptsize{± 0.158} & 0.098 \scriptsize{± 0.000}   
        & 2.586 \scriptsize{± 0.000} & 1.938 \scriptsize{± 0.000} & 0.327 \scriptsize{± 0.000}\\
& 36    & \textbf{5.111} \scriptsize{± 0.001} & 2.474 \scriptsize{± 0.022} & 0.288 \scriptsize{± 0.015}                       
        & 4.724 \scriptsize{± 0.046} & 1.516 \scriptsize{± 0.132} & 0.125 \scriptsize{± 0.019} 
        & 4.662 \scriptsize{± 0.042} & 2.393 \scriptsize{± 0.016} & 0.349 \scriptsize{± 0.036} \\
& 48    & 7.335 \scriptsize{± 0.037} & \underline{2.810} \scriptsize{± 0.029} & \underline{0.287} \scriptsize{± 0.009}     
        & 8.245 \scriptsize{± 0.407} & 1.760 \scriptsize{± 0.041} & 0.129 \scriptsize{± 0.001}   
        & 6.955 \scriptsize{± 0.507} & 2.855 \scriptsize{± 0.067} & 0.415 \scriptsize{± 0.077}\\
& 60    & 12.14 \scriptsize{± 0.066} & 3.668 \scriptsize{± 0.102} & 0.380 \scriptsize{± 0.071}           
        & 10.53 \scriptsize{± 0.132} & \textbf{2.026} \scriptsize{± 0.022} & 0.139 \scriptsize{± 0.002}    
        & 9.944 \scriptsize{± 0.132} & 2.854 \scriptsize{± 0.132} & 0.415 \scriptsize{± 0.132}\\\midrule
\multirow{4}{*}{\begin{sideways}FEDformer\end{sideways}}  
& 24    & 4.335 \scriptsize{± 0.021} & 1.754 \scriptsize{± 0.051} & 0.243 \scriptsize{± 0.009}                         
        & 3.311 \scriptsize{± 0.064} & 0.887 \scriptsize{± 0.036} & 0.079 \scriptsize{± 0.006} 
        & 2.872 \scriptsize{± 0.027} & \underline{1.506} \scriptsize{± 0.058} & 0.215 \scriptsize{± 0.007}\\
& 36    & 6.794 \scriptsize{± 0.022} & 2.505 \scriptsize{± 0.063} & 0.304 \scriptsize{± 0.010}            
        & 5.638 \scriptsize{± 0.093} & 1.079 \scriptsize{± 0.014} & 0.085 \scriptsize{± 0.009} 
        & 5.108 \scriptsize{± 0.023} & 1.801 \scriptsize{± 0.039} & 0.313 \scriptsize{± 0.003}\\
& 48    & 8.203 \scriptsize{± 0.033} & 2.891 \scriptsize{± 0.101} & 0.308 \scriptsize{± 0.000}            
        & 7.952 \scriptsize{± 0.077} & 1.692 \scriptsize{± 0.019} & 0.108 \scriptsize{± 0.002} 
        & \underline{6.342} \scriptsize{± 0.096} & 2.053 \scriptsize{± 0.033} & 0.226 \scriptsize{± 0.004}\\
& 60    & 12.76 \scriptsize{± 0.054} & \textbf{3.209} \scriptsize{± 0.091} & \underline{0.321} \scriptsize{± 0.000}
        & 10.68 \scriptsize{± 0.226} & 2.714 \scriptsize{± 0.054} & 0.128 \scriptsize{± 0.009} 
        & \underline{9.495} \scriptsize{± 0.132} & 2.083 \scriptsize{± 0.132} & \underline{0.199} \scriptsize{± 0.132}\\\midrule
\multirow{4}{*}{\begin{sideways}PatchTST\end{sideways}}     
& 24    & 4.222 \scriptsize{± 0.098} & 1.529 \scriptsize{± 0.102} & 0.215 \scriptsize{± 0.009}            
        & 3.658 \scriptsize{± 0.138} & 0.903 \scriptsize{± 0.122} & 0.056 \scriptsize{± 0.064} 
        & 3.089 \scriptsize{± 0.132} & 1.796 \scriptsize{± 0.132} & \underline{0.119} \scriptsize{± 0.132}\\
& 36    & 6.388 \scriptsize{± 0.094} & \textbf{2.154} \scriptsize{± 0.099} & \textbf{0.234} \scriptsize{± 0.007}      
        & 5.603 \scriptsize{± 0.436} & \textbf{0.776} \scriptsize{± 0.026} & 0.078 \scriptsize{± 0.034} 
        & 4.849 \scriptsize{± 0.436} & \underline{1.428} \scriptsize{± 0.436} & 0.149 \scriptsize{± 0.036} \\
& 48    & 10.49 \scriptsize{± 0.090} & 3.075 \scriptsize{± 0.110} & 0.356 \scriptsize{± 0.088}                       
        & 8.083 \scriptsize{± 0.647} & 1.701 \scriptsize{± 0.027} & 0.099 \scriptsize{± 0.046} 
        & 6.687 \scriptsize{± 0.647} & \textbf{1.473} \scriptsize{± 0.647} & 0.181 \scriptsize{± 0.017}\\
& 60    & 14.09 \scriptsize{± 0.088} & 3.883 \scriptsize{± 0.098} & 0.693 \scriptsize{± 0.009}                       
        & 11.32 \scriptsize{± 0.169} & 2.210 \scriptsize{± 0.016} & \textbf{0.106} \scriptsize{± 0.006} 
        & 10.40 \scriptsize{± 0.049} & \underline{1.988} \scriptsize{± 0.029} & 0.229 \scriptsize{± 0.169} \\\bottomrule
\multirow{4}{*}{\begin{sideways}CONTIME\end{sideways}} 
& 24    & \textbf{4.218} \scriptsize{± 0.021} & \underline{1.528} \scriptsize{± 0.041} & \textbf{0.184} \scriptsize{± 0.002} 
        & \textbf{1.761} \scriptsize{± 0.027} & \underline{0.884} \scriptsize{± 0.013} & \underline{0.049} \scriptsize{± 0.007} 
        & \textbf{2.254} \scriptsize{± 0.011} & \textbf{1.023} \scriptsize{± 0.021} & \textbf{0.117} \scriptsize{± 0.002} \\
& 36    & \underline{5.371} \scriptsize{± 0.018} & \underline{2.334} \scriptsize{± 0.007} & \underline{0.256} \scriptsize{± 0.003} 
        & \textbf{3.488} \scriptsize{± 0.017} & 1.221 \scriptsize{± 0.007} & \textbf{0.063} \scriptsize{± 0.009} 
        & \textbf{4.120} \scriptsize{± 0.015} & \textbf{1.390} \scriptsize{± 0.012} & \textbf{0.136} \scriptsize{± 0.003} \\
& 48    & \textbf{7.296} \scriptsize{± 0.024} & \textbf{2.755} \scriptsize{± 0.010} & \textbf{0.262} \scriptsize{± 0.005} 
        & \textbf{5.366} \scriptsize{± 0.037} & \textbf{1.683} \scriptsize{± 0.071} & \underline{0.097} \scriptsize{± 0.011} 
        & \textbf{6.226} \scriptsize{± 0.033} & \underline{1.805} \scriptsize{± 0.010} & \textbf{0.159} \scriptsize{± 0.005} \\
& 60    & \textbf{11.83} \scriptsize{± 0.041} & \underline{3.261} \scriptsize{± 0.009} & \textbf{0.292} \scriptsize{± 0.007}
        & \textbf{7.452} \scriptsize{± 0.080} & \underline{2.139} \scriptsize{± 0.001} & 0.125 \scriptsize{± 0.004}
        & \textbf{9.366} \scriptsize{± 0.019} & 2.121 \scriptsize{± 0.091} & \textbf{0.174} \scriptsize{± 0.002} \\\bottomrule
\end{tabular}
\end{table*}

\end{document}